\documentclass[10pt,twocolumn,letterpaper]{article}

\usepackage{iccv}
\usepackage{times}
\usepackage{epsfig}
\usepackage{graphicx}
\usepackage{amsmath}
\usepackage{amssymb}
\usepackage{booktabs}
\usepackage{multirow, makecell}   
\usepackage{color, colortbl}   
\usepackage[table]{xcolor}
\usepackage{caption} 
\usepackage{standalone}

\usepackage[pagebackref=true,breaklinks=true,letterpaper=true,colorlinks,bookmarks=false]{hyperref}

\iccvfinalcopy 

\def\iccvPaperID{5210} 

\ificcvfinal\fi

\begin{document}

\title{Improving Scene Text Recognition for Character-Level Long-Tailed Distribution}






%

\author{
Sunghyun Park\textsuperscript{*}\textsuperscript{\rm 1}\quad\quad
Sunghyo Chung\textsuperscript{*}\textsuperscript{\rm 2}\quad\quad
Jungsoo Lee\textsuperscript{\rm 1}\quad\quad
Jaegul Choo\textsuperscript{\rm 1}\\ \\
\textsuperscript{\rm 1} KAIST\quad\quad
\textsuperscript{\rm 2} Kakao Enterprise\quad\quad\\
{\tt\small \{psh01087, bebeto, jchoo\}@kaist.ac.kr}\quad\quad 
{\tt\small shawn.c@kakaoenterprise.com} \\ \\
}

\newcommand\blfootnote[1]{
  \begingroup
  \renewcommand\thefootnote{}\footnote{#1}
  \addtocounter{footnote}{-1}
  \endgroup
}

\blfootnote{\textsuperscript{*} These authors contributed equally. \vspace{-0.7cm}}

\ificcvfinal\thispagestyle{empty}\fi

\begin{figure}
\twocolumn[{
\renewcommand\twocolumn[1][]{#1}
\maketitle
\vspace{-1.1cm}
\begin{center}
    \centering 
    \includegraphics[width=1.0\linewidth]{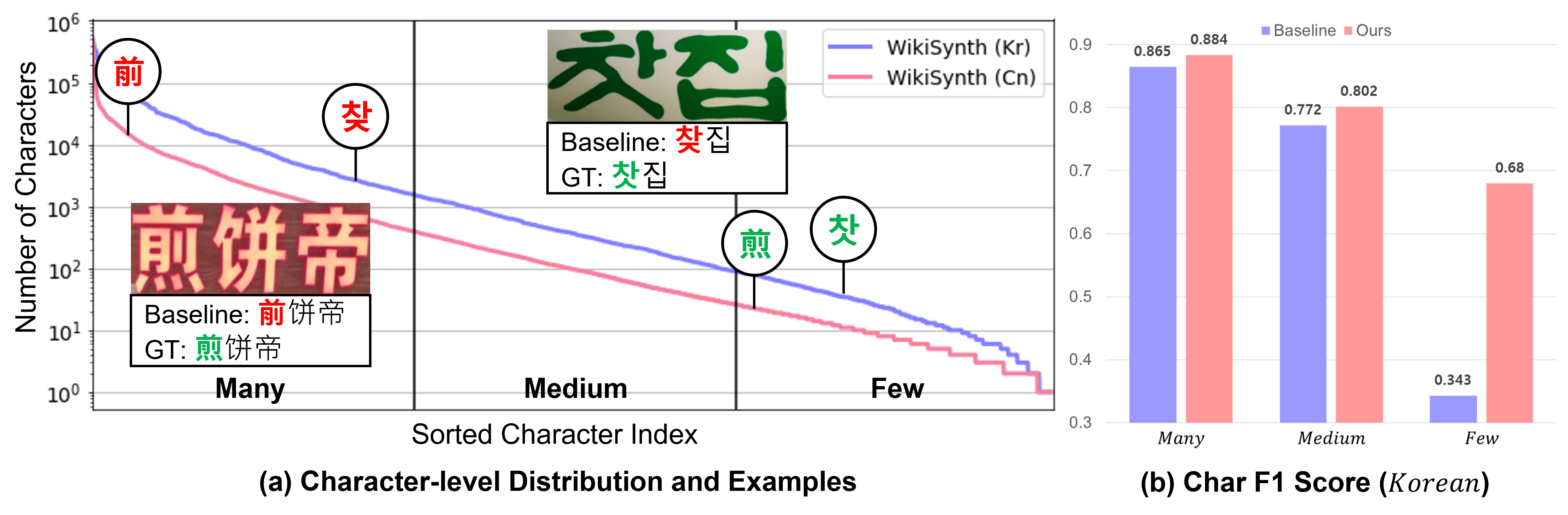}
    \vspace{-0.8cm}
    \captionof{figure}{(a) We visualize the character-level distributions of WikiSynth of Korean (Kr) and Chinese (Cn). 
    We categorize the characters according to the number of training samples: \emph{many}, \emph{medium}, and \emph{few}.
    We also show misclassified images of tail characters predicted wrongly as head characters.
    (b) Our approach outperforms the baseline model when evaluated with character-level (char) F1 score, a newly proposed evaluation metric, which measures the performance at the character level. The higher score, the better.
    This result shows that our method enhances the performance on \emph{few} characters significantly.}
    \vspace{-0.1cm}
    \label{fig:main}
\end{center}
}]
\end{figure}

\begin{abstract}
Despite the recent remarkable improvements in scene text recognition (STR), the majority of the studies focused mainly on the English language, which only includes few number of characters.
However, STR models show a large performance degradation on languages with a numerous number of characters (e.g., Chinese and Korean), especially on characters that rarely appear due to the long-tailed distribution of characters in such languages.
To address such an issue, we conducted an empirical analysis using synthetic datasets with different character-level distributions (e.g., balanced and long-tailed distributions). 
While increasing a substantial number of tail classes without considering the context helps the model to correctly recognize characters individually, training with such a synthetic dataset interferes the model with learning the contextual information (i.e., relation among characters), which is also important for predicting the whole word.
Based on this motivation, we propose a novel Context-Aware and Free Experts Network (CAFE-Net) using two experts: 1) context-aware expert learns the contextual representation trained with a long-tailed dataset composed of common words used in everyday life and 2) context-free expert focuses on correctly predicting individual characters by utilizing a dataset with a balanced number of characters.
By training two experts to focus on learning contextual and visual representations, respectively, we propose a novel confidence ensemble method to compensate the limitation of each expert. 
Through the experiments, we demonstrate that CAFE-Net improves the STR performance on languages containing numerous number of characters.
Moreover, we show that CAFE-Net is easily applicable to various STR models.
\end{abstract}

\section{Introduction}

Recent studies in scene text recognition (STR) models have shown remarkable performances. 
As the most commonly spoken language worldwide, the English language has been the main focus of the existing STR studies~\cite{shi2018aster,baek2019STRcomparisons,baek2021STRfewerlabels,wan2020vocabulary}.
However, achieving high performance on other languages with the existing models is a non-trivial task, especially when the languages have numerous characters (\textit{e.g.,} letter, number, symbol.), unlike English.
More specifically, English has only 26 letters, while Asian languages like Chinese and Korean have thousands of letters.

There exist few studies that try to improve STR performance on languages other than English~\cite{buvsta2018e2e,huang2021multiplexed}.
However, they overlook the fact that languages with a large number of characters have the \emph{long-tailed distribution at the character level}.
Due to the character-level long-tailed distribution, the model mainly focuses on learning the head characters (\textit{i.e.,} those which frequently appear when forming words) while focusing less on learning the tail characters (\textit{i.e.,} those which rarely appear in words).
This leads to significant performance degradation on the tail classes, a commonly observed phenomenon in the existing long-tailed recognition~\cite{kang2019decoupling,ren2020balanced}, as shown in Fig.~\ref{fig:main}.

Although synthetic datasets such as SynthText~\cite{Gupta16} are often utilized in STR, constructing synthetic datasets with a balanced number of characters is challenging.
To be more specific, to alleviate the performance degradation due to the long-tailed distribution, the existing image classification methods studies generally proliferate the data samples of the tail classes when constructing a balanced set of classes. 
In STR, however, increasing the number of words including the tail characters also increases the number of the head characters when they are included in the same word.
While generating words only including tail classes is one straightforward solution, those words generally do not include the contexts people use in their everyday life since tail classes are rarely used in common words. 
Such an issue makes it demanding to construct a synthetic dataset for STR that can improve the performance on the tail characters, especially when the characters show a long-tailed distribution.

This paper is the first work to address the STR task in terms of the \emph{character-level long-tailed distribution}.
Such a long-tailed distribution of the characters causes a significant performance drop on tail characters in STR.
We investigate the character-level long-tailed distribution by constructing two synthetic datasets having different character-level distributions: (1) one created by common words to preserve the context (\textit{i.e.,} WikiSynth) and (2) the other with randomly combined characters, which has a balanced distribution but lacks such contextual information (\textit{i.e.,} RandomSynth).
While training with WikiSynth encourages the model to learn contextual information, the model fails to predict the tail classes correctly due to the long-tailed distribution of characters. 
In contrast, using RandomSynth helps to correctly predict characters individually by focusing on the visual representation and enhances the performance on tail classes significantly, but such training interferes the model with learning the contextual information.

Based on the findings, we propose a Context-Aware and Free Experts Network (CAFE-Net), a \textit{simple yet effective} approach, which utilizes the confidence score for aggregating experts handling different character-level distributions.
At a high level, we train two experts separately: (1) context-aware expert that focuses on learning the contextual representation using a dataset including characters with a long-tailed distribution and (2) context-free expert that learns the visual representation by utilizing a dataset of a balanced number of characters. 
Additionally, we propose a new evaluation metric termed `character-level (char) F1 score', which is more suitable than existing word-level evaluation metrics (\textit{e.g.}, accuracy) for character-level analysis.
Extensive experiments demonstrate that CAFE-Net significantly outperforms the existing methods in predicting the tail characters, while improving the performance on predicting the whole words with languages containing a numerous number of characters.
Furthermore, we demonstrate the applicability of CAFE-Net by combining various STR models.

The main contributions of our work are as follows:
\begin{itemize}
    \item To the best of our knowledge, this is the first work to handle the STR model in terms of the languages with character-level long-tailed distributions.
    \item To take care of learning both contextual and visual information, we propose a novel CAFE-Net using context-aware and context-free experts, which separately handles different character-level distributions.
    \item We demonstrate the superior performance of our method and its applicability through experiments.
\end{itemize}

\section{Related Work}

\noindent\textbf{Scene Text Recognition.}
A recent study~\cite{baek2019STRcomparisons} proposes a widely used STR framework composed of four stages by analyzing the performance gains of each module in a model.
Leveraging such a well-performing framework, the majority of studies in STR mainly focused on English~\cite{mou2020plugnet,yue2020robustscanner}. 
Several studies~\cite{buvsta2018e2e,huang2021multiplexed} propose unified methods to handle multiple languages.
However, such existing multilingual STR approaches do not consider the characteristics of each language (\textit{e.g.,} long-tailed distribution of characters).
Another recent work tackled the vocabulary reliance problem~\cite{wan2020vocabulary} at the word level, which mitigates the poor generalization on images with words not included in the vocabulary of a training dataset.
In contrast to the previous STR studies, to the best of our knowledge, this is the first work to address the \emph{character level} long-tailed distribution in STR.

\noindent\textbf{Long-tailed Recognition.}
There exist numerous datasets, which have long-tailed distributions in the real world. 
Previous studies addressing the long-tailed distribution focused on improving loss functions~\cite{lin2017focal,cui2019class}, augmenting the data samples of the tail classes~\cite{chawla2002smote,li2021metasaug}, and adjusting the logits~\cite{kang2019decoupling, lee2021improving, menon2020long,ren2020balanced}.
Recent studies proposed using multiple experts specialized for correctly predicting under a certain label distribution~\cite{zhou2020bbn,wang2020long,zhang2021test}.
Such a design enables to handle different label distributions within a single model.
Inspired by such a design, we train two different experts specialized to learn contextual and visual representation, respectively, by taking account of the characteristic of STR.

\section{Motivation}\label{sec:investigating}

\noindent\textbf{Overview.}
This section investigates the impacts of character-level long-tailed distribution in STR.
We first describe several synthetic datasets, which are generated by shifting the character-level distribution (\textit{e.g.}, varying from long-tailed datasets to balanced datasets) in Section.~\ref{sec:synthetic data}.
Moreover, we introduce character-level F1 score in Section.~\ref{sec:char-f1}.
Next, we show the effectiveness of each synthetic dataset and analyze them in Section.~\ref{sec:tradeoff}.
We use a TRBA model~\cite{baek2019STRcomparisons}, a representative STR framework, for the experiments in this section.
The details for STR framework we used are described in the supplementary.


\begin{figure}[t!]
    \centering
    \includegraphics[width=1.0\linewidth]{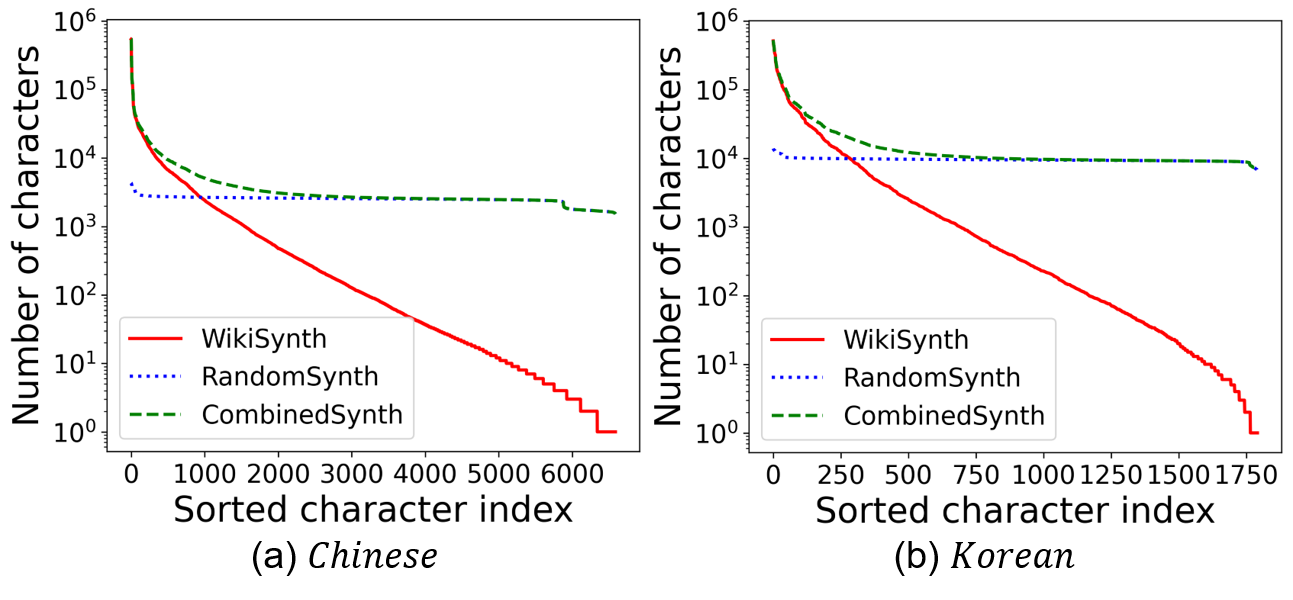}
    \vspace{-0.7cm}
    \caption{Character-level distribution of WikiSynth (WS), RandomSynth (RS), and CombinedSynth (CS). Unlike WS, both RS and CS include a sufficient number of characters for all classes.}
    \vspace{-0.5cm}
    \label{fig:distribution}
\end{figure}

\subsection{Synthetic Data}\label{sec:synthetic data}
As widely used in the previous studies of STR~\cite{shi2018aster,baek2019STRcomparisons,yu2020towards,fang2021read,atienza2021vision}, we utilize synthetic data for training.
We use Korean and Chinese for the languages, which include the long-tailed distributions at the character level. 
We construct the training datasets for each language by following SynthText~\cite{Gupta16}, which is one of the synthetic datasets generated by using a large corpus and diverse backgrounds.
We generate new synthetic datasets for the study by shifting the character-level distribution as shown in Fig.~\ref{fig:distribution}.

\begin{figure}[t!]
    \centering
    \includegraphics[width=1.0\linewidth]{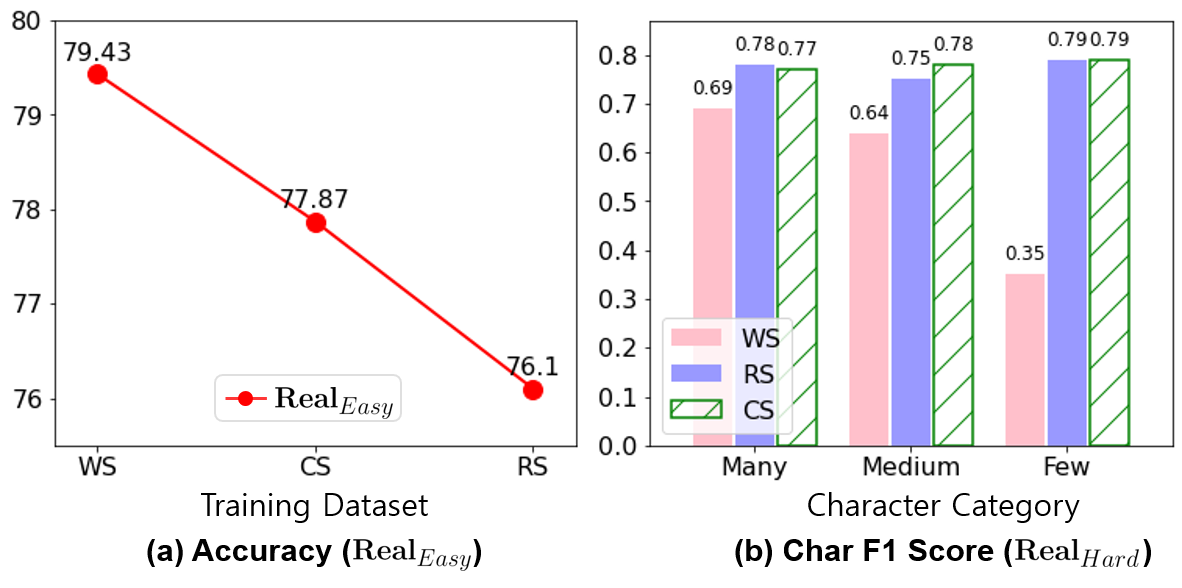}
    \vspace{-0.7cm}
    \caption{(a) Accuracy on \textbf{Real}$_{Easy}$ of models trained with WS, CS, and RS individually using Korean. Since RS and CS include randomly combining characters, the models trained with RS and CS exhibit lower accuracy compared to the model trained on WS. (b) Char F1 score on \textbf{Real}$_{Hard}$.
    We observe that training models with RS and CS improve the recognition performance on individual characters.}
    \vspace{-0.6cm}
    \label{fig:tradeoff dataset}
\end{figure}

\noindent\textbf{WikiSynth (WS)}
This dataset utilizes Wikipedia text corpus.
The wiki corpus is composed of word units using a tokenizer for each language.
The limit of word length is set to 25. 
The number of samples in the training and test sets for Chinese and Korean are 5,000,000 and 10,000, respectively.
Since WS is generated by common words, it has a long-tailed distribution at character level that is generally observed in languages with numerous number of characters. 

\noindent\textbf{RandomSynth (RS)}
In contrast to WS, RS is a character-level balanced dataset, where words are generated by randomly combining characters.
Since RS samples the characters uniformly, the dataset does not consider the context, so it does not contain the words generally used in the real world.
RS contains the same number of images as WS for a fair comparison.
As previous studies in long-tailed recognition~\cite{cao2019learning} evaluate the models with the balanced test set, we use RS as the character-level balanced test set in STR. 

\noindent\textbf{CombinedSynth (CS)}
WS and RS has each own limitation, respectively.
To be more specific, models trained with WS fail to learn \emph{few} characters, while training with RS interferes the model with learning the contextual information between characters.
A viable option for solving these problems is to mix WS and RS.
CS is composed of WS and RS with an equal number of images from each dataset to compensate for the limitation of each dataset.

\subsection{Character-Level F1 Score}\label{sec:char-f1}
Accuracy is a widely used evaluation metric, which evaluates whether a model correctly outputs all the characters in a given word.
Since the accuracy only considers the performance of STR at the word level, we propose a novel evaluation metric termed `char F1 score' to evaluate the performance on the character level. 
When obtaining the char F1 score, we 1) perform the sequence alignment of ground truth and predicted characters, 2) compute the F1 score per character, and 3) average these scores.
We report the F1 score in addition to the accuracy since it is more suitable than accuracy when evaluating models with an imbalanced number of data samples.
The details of char F1 score are described in the supplementary.

Since we address the long-tailed distribution of characters, we categorize the characters into three groups.
For simplicity, we denote $n_i$ as the number of training samples including $i^{\text{th}}$ character in a given dataset.
The characters are categorized according to $n_i$: 1) \emph{many} (\textit{i.e.,} $n_i \geq 1500$), 2) \emph{medium} (\textit{i.e.,} $n_i<1500$ and $n_i \geq 100$), and 3) \emph{few} (\textit{i.e.,} $n_i<100$).
Straightforwardly, char F1 scores of \emph{few} characters are much lower than those of \emph{many} characters when training models with WS as shown in Fig.~\ref{fig:tradeoff dataset} (b).

\subsection{Tradeoff between Context-Free and Context-Aware Learning}
\label{sec:tradeoff}

\begin{figure*}[t!]
    \centering
    \includegraphics[width=1.0\linewidth]{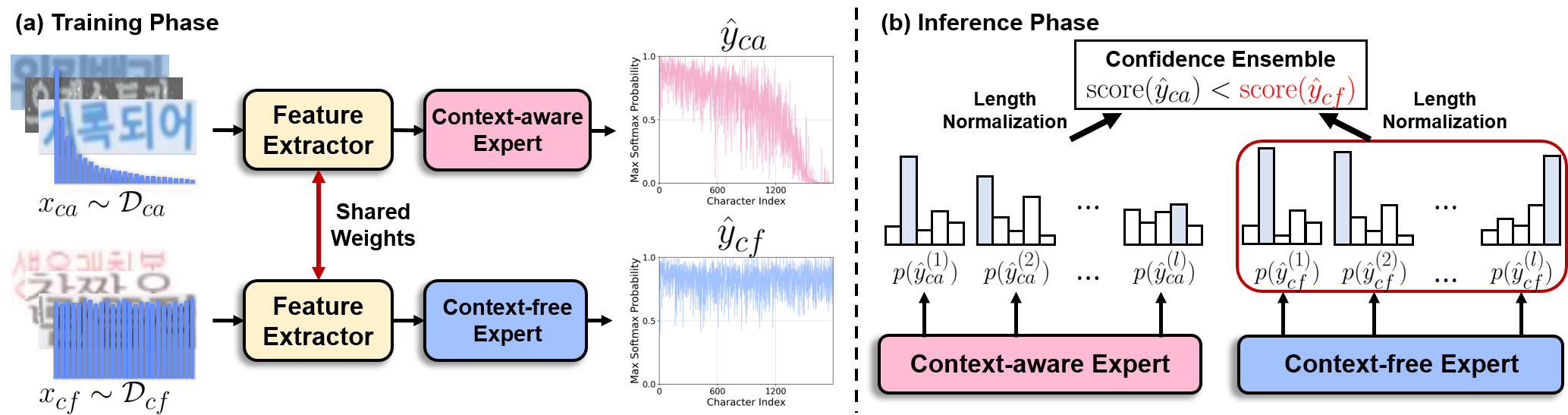}
    \vspace{-0.8cm}
    \caption{Overview of CAFE-Net. CAFE-Net utilizes two different experts: a context-aware expert and a context-free expert. Context-aware expert is trained with $\mathcal{D}_{ca}$ (\textit{e.g.}, WS) to focus on learning the contextual information. On the other hand, context-free expert focuses on learning recognizing individual characters, which is trained with $\mathcal{D}_{cf}$ (\textit{e.g.}, RS), a balanced dataset with images including randomly sequenced characters. 
    As evidenced by visualization of the maximum softmax probabilities of two experts, it is clear that the experts have different certainty depending on the characters.
    Based on this characteristic, CAFE-Net select the prediction with the higher confidence score from the two experts during the inference.
    }
    \vspace{-0.5cm}
    \label{fig:model overview}
\end{figure*}

We use AI Hub dataset~\cite{aihub}, a publicly available Korean dataset, for Korean test set noted as `\textbf{Real}'.
Additionally, we divide \textbf{Real} datasets into two types of test sets: 1) a test set without \emph{few} characters (\textit{i.e.}, \textbf{Real}$_{Easy}$) and 2) a test set including \emph{few} characters (\textit{i.e.}, \textbf{Real}$_{Hard}$).
The details of the experimental setup are described in the supplementary.

We evaluate the models with \textbf{Real}$_{easy}$ and \textbf{Real}$_{hard}$ by individually training them with WS, RS, and CS using Korean.
Note that the model trained with WS mainly primarily relies on contextual information for making predictions, whereas the one trained with RS mainly uses visual information while lacking contextual information.
We observe a tradeoff of using WS and RS for the training set. 
Fig.~\ref{fig:tradeoff dataset} (a) demonstrates that training with WS improves the \textit{accuracy} on \textbf{Real}$_{easy}$ compared to training with CS or RS.
On the other hand, Fig.~\ref{fig:tradeoff dataset} (b) shows that training with CS or RS improves the \textit{char F1 score} for all \textit{many}, \textit{medium}, \textit{few} characters when evaluated with \textbf{Real}$_{easy}$ compared to training with WS.

Through the experiments, we found that the model focused on learning visual information without contexts (\textit{i.e.}, trained with RS or CS) can correctly predict individual characters, which is important for improving the performance of long-tailed recognition, especially for \textit{few} characters.
However, the model focusing on learning the contextual information (\textit{i.e.,} trained with WS) shows improved accuracy even with low char F1 score.
This indicates that capturing the contextual information is crucial for correctly predicting all characters of a given word, especially for those words frequently appearing.
Without such understanding of the contextual information, models show limited accuracy with even high char F1 score. 
Therefore, to improve recognizing individual characters and the whole word, we need to enhance both visual and contextual representations. 


\section{Method}\label{sec:method}

\noindent\textbf{Overview.}
Based on the empirical analysis, we propose a Context-Aware and Free Experts Network termed `CAFE-Net'.
Different from previous STR methods, we utilize two types of training datasets, which have different label distributions (\textit{e.g.}, WS and RS).
As described in Fig.~\ref{fig:model overview}, our model consists of two main experts: (1) context-aware expert trained with WS to focus on the contextual representation via utilizing an external language model; (2) context-free expert trained with a balanced number of characters (\textit{i.e.}, RS) to improve the performance on \emph{few} characters.
By dividing the roles of two experts, it is possible to improve the performance on \emph{few} characters while understanding the contextual information.

Different from the existing STR methods, we utilize two synthetic datasets (\textit{i.e.}, WS and RS) separately during training.
Let $\{ x_{ca}, y_{ca} \} \sim \mathcal{D}_{ca}$ and $\{ x_{cf}, y_{cf} \} \sim \mathcal{D}_{cf}$ denote training images and labels sampled for training the context-aware expert and the context-free expert, respectively.
In specific, we utilize WS and RS for $\mathcal{D}_{ca}$ and $\mathcal{D}_{cf}$, respectively.
In the following, we illustrate the details of our method and its objective functions.


\noindent\textbf{Feature Extractor.}
$x_{ca}$ and $x_{cf}$ are fed into the feature extractor to acquire the context-aware and context-free feature representations $f_{ca}$ and $f_{cf}$, respectively.
In our framework, two experts share the same feature extractor.
Sharing weights largely reduces the computational complexity in the inference phase.
For the feature extractor, various model architectures can be utilized such as ResNet~\cite{he2016deep} and vision transformer (ViT) encoder~\cite{atienza2021vision}.

\noindent\textbf{Context-Free Expert.}
Given the feature representation $f_{cf}$ that is extracted from $x_{cf}$, a context-free expert produces the output feature $h_{cf}=\{ h^{(1)}_{cf}, \dots, h^{(T)}_{cf} \}$ of the corresponding words $\hat{y}_{cf}=\{ \hat{y}^{(1)}_{cf}, \dots, \hat{y}^{(T)}_{cf} \}$.
Here, $T$ denotes the maximum length of the word.
Due to the balanced number of characters, the context-free expert correctly predicts \emph{few} characters more compared to the context-aware expert.
This is mainly due to the fact that the random sequences of characters devoid of semantic meaning make the context-free expert prioritize learning visual representation above contextual representation.

\noindent\textbf{Context-Aware Expert.}
Different from the context-free expert, the context-aware expert is trained with $\mathcal{D}_{ca}$ to focus on learning the contextual information, which is essential to predict the whole words accurately.
Inspired by recent context-aware STR methods~\cite{yu2020towards,fang2021read}, we leverage an external language model to capture semantic information to assist STR.
Specifically, with the feature representations $f_{ca}$ and $f_{cf}$, the context-aware expert produces the output feature.
Then, an external language model refines the output of the context-aware expert.
Finally, the outputs of the context-aware expert and the language model are fused to produce the final output feature.
In summary, the context-aware expert with the external language model produces the final output feature $h_{ca}=\{ h^{(1)}_{ca}, \dots, h^{(T)}_{ca} \}$ of the corresponding words $\hat{y}_{ca}=\{ \hat{y}^{(1)}_{ca}, \dots, \hat{y}^{(T)}_{ca} \}$.

\noindent\textbf{Objective Functions.}
The context-free and context-aware experts are trained by the same objective function that minimizes negative log-likelihood of the conditional probability of word label $y_{cf}$.
Formally, loss function $\mathcal{L}$ is as follows:
\begin{equation}
    \mathcal{L} = -\frac{1}{T} \sum^{T}_{t=1} \log p(y^t|h^t),
\end{equation}
where $y^t$ is the $t$-th ground truth character.

\noindent\textbf{Confidence Ensemble.}
During inference, we aggregate the outputs of two experts.
The output probability of each expert is defined as:
\begin{equation}
    p(\hat{y}) = \prod^{l}_{t=1} p(\hat{y}^t|\hat{y}^{<t}),
\end{equation}
where $\hat{y}^{<t} = y^1 \dotsm y^{t-1}$ and $l$ is the length of the predicted words.
In specific, we ignore $pad$ token and only consider the words preceding the $eos$ token, where $eos$ token indicates the end of the words.

\begin{table*}[t!]
    \small
    \centering
    \begin{tabular}{@{}c@{}c|p{0.07\textwidth}<{\centering}p{0.07\textwidth}<{\centering}p{0.07\textwidth}<{\centering}p{0.07\textwidth}<{\centering}|p{0.07\textwidth}<{\centering}p{0.07\textwidth}<{\centering}p{0.07\textwidth}<{\centering}p{0.07\textwidth}<{\centering}@{}}
        \toprule
        \multirowcell{2}{\multicolumn{1}{c}{\textbf{Method.}}} & 
        \multirowcell{2}{\textbf{Train} \\ \textbf{Data}} & 
        \multicolumn{4}{c|}{\textit{Korean}} & \multicolumn{4}{c}{\textit{Chinese}} \\
        && \textbf{Real} & \textbf{Real}$_{Easy}$ & \textbf{Real}$_{Hard}$ & \textbf{Synth}$_{test}$
        & \textbf{Real} & \textbf{Real}$_{Easy}$ & \textbf{Real}$_{Hard}$ & \textbf{Synth}$_{test}$ \\
        \midrule
        \textbf{CNN-based} \\
        \midrule
        \multirowcell{2}{TRBA}
         & WS & 78.25 & 79.43 & 34.14 & 87.47
              & 39.19 & 42.99 & 9.25 & 83.23 \\
         & CS & 77.43 & 77.87 & 61.25 & 86.37 
              & 41.83 & 41.72 & 42.71 & 83.31 \\
         \rowcolor{gray!10} 
         +Ours & CS & \textbf{81.35} & \textbf{81.75} & \textbf{66.68} & \textbf{88.93} 
                    & \textbf{47.67} & \textbf{48.09} & \textbf{44.34} & \textbf{86.22} \\
        \midrule
        \multirowcell{2}{TextAdaIN}
         & WS & 80.35 & 81.57 & 34.54 & 86.97 
              & 41.33 & 45.78 & 6.30 & 81.73 \\
         & CS & 80.43 & 80.80 & 66.60 & 85.82
              & 45.76 & 45.82 & \textbf{45.30} & 81.80 \\
         \rowcolor{gray!10} 
         +Ours & CS & \textbf{82.34} & \textbf{82.75} & \textbf{66.85} & \textbf{88.88} 
                    & \textbf{47.21} & \textbf{47.85} & 42.15 & \textbf{85.83} \\
        \midrule
        \textbf{ViT-based} \\
        \midrule
        \multirowcell{2}{ViTSTR + Linear}
         & WS & 80.92 & 81.74 & 50.46 & 90.57
              & 44.14 & 46.12 & 28.51 & 89.81 \\
         & CS & 81.82 & 82.19 & 68.07 & 90.93 
              & 49.15 & 48.84 & 51.63 & 90.51 \\
         \rowcolor{gray!10} 
         +Ours & CS & \textbf{82.78} & \textbf{83.14} & \textbf{69.16} & \textbf{92.09}
                    & \textbf{51.37} & \textbf{51.10} & \textbf{53.47} & \textbf{91.09} \\
        \midrule
        \multirowcell{2}{ViTSTR + Attn}
         & WS & 83.39 & 84.05 & 58.90 & 91.24 
              & 48.22 & 50.39 & 31.17 & 89.35 \\
         & CS & 83.56 & 83.91 & 70.38 & 90.82 
              & 50.94 & 51.14 & 49.34 & 89.27 \\
         \rowcolor{gray!10} 
         +Ours & CS & \textbf{85.39} & \textbf{85.75} & \textbf{72.17} & \textbf{91.66}
                    & \textbf{55.21} & \textbf{55.45} & \textbf{53.38} & \textbf{91.24} \\
        \bottomrule
    \end{tabular}
    \vspace{-0.3cm}
    \caption{Accuracy on Korean and Chinese datasets. \textit{2nd} column indicates training synthetic datasets. Applying CAFE-Net consistently improves the performance on various evaluation datasets: \textbf{Real}, \textbf{Real}$_{Easy}$, \textbf{Real}$_{Hard}$ and \textbf{Synth}$_{test}$.}
    \vspace{-0.5cm}
    \label{Table:experiment-ours}
\end{table*}

To ensemble the outputs of two experts, we leverage the maximum softmax probability, which represents the probability $p(\hat{y}^t|\hat{y}^{<t})$ of the predicted character.
The confidence score $\text{score}(\hat{y})$ of each expert is calculated based on the maximum softmax probability of the characters as follows:
\begin{equation}
    \text{score}(\hat{y}) = \frac{1}{l} \sum^{l}_{t=1} \log ( \operatorname*{max}(p(\hat{y}^t|\hat{y}^{<t})) ),
\end{equation}
where we apply the length normalization that normalizes the score using the length $l$ of the predicted word.
Since the probabilities $p(\hat{y}^t)$ are all values less than one, multiplying a non-trivial number of values less than one will result in the confidence score of shorter words increasing.
To address this issue, we normalize the confidence score by dividing it by the word length $l$.
We denote the confidence scores of the context-aware expert and context-free expert as $\text{score}(\hat{y}_{ca})$ and $\text{score}(\hat{y}_{cf})$, respectively.
Among $\text{score}(\hat{y}_{ca})$ and $\text{score}(\hat{y}_{cf})$, we select the output with the higher confidence score.
Then, the final predicted words are computed by taking the highest probability character at each time step $t$.
Intuitively, since the maximum softmax probabilities of the two experts vary depending on the characters, CAFE-Net is capable of selecting the word prediction properly during inference by utilizing the confidence score obtained from the two experts.

\noindent\textbf{Applicability of CAFE-Net.}
Our proposed method provides a practical solution for addressing character-level long-tailed distribution in various STR models.
In the supplementary, we describe how to integrate our method with representative STR models such as CNN-based models~\cite{baek2019STRcomparisons,nuriel2022textadain} and ViT-based models~\cite{atienza2021vision}.
While ensembling or utilizing multiple experts has been widely explored in other fields~\cite{biasensemble,wang2020long,zhou2020bbn,zhang2021test}, we want to emphasize that we shed light on \textit{how} to utilize ensembling in the character-level long-tailed STR. 
Notably, the key difference between character-level long-tailed STR and previous studies is that STR includes both vision and language modalities, where the model requires both visual and contextual information to predict the whole words. 
Due to this fact, simply adopting previous ensembling methods may not be directly applicable in STR. 
To solve such an issue, we first discover a crucial finding and propose a \textit{simple yet effective} method based on our finding.

\section{Experiments}

\noindent\textbf{Experimental Setup.}
Since we only use synthetic datasets for Chinese and Korean, we also utilize ICDAR 2017 MLT dataset (MLT17)~\cite{nayef2017icdar2017}, a real-world dataset, for each language to reduce the domain gap with the real-world datasets.
We filter the images of MLT17 including Chinese and Korean for each language.
We evaluate the model using accuracy, a widely used evaluation metric in STR.

We evaluate the performance of the models on large-scale real-world datasets.
We utilize real-world datasets as test sets noted as `\textbf{Real}'.
In specific, AI Hub dataset~\cite{aihub} and ICDAR 2019 ReCTS dataset~\cite{sun2019icdar} are publicly available real-world Korean and Chinese datasets, respectively.
AI Hub dataset, a Korean real STR dataset, includes 151,105 cropped images of street signs, traffic signs, and brands.
ReCTS, a Chinese real STR dataset, contains 77,709 cropped images of Chinese signboards in the street view with diverse backgrounds and fonts, which is a widely used benchmark dataset in the STR field.
We choose these two datasets for evaluation since they contain a sufficient number of tail characters.
The details for preprocessing real-world datasets are depicted in the supplementary.

We assess the performance of the models using the synthetic test datasets (\textit{e.g.}, WS$_{test}$ and RS$_{test}$) in addition to real-world datasets.
WS$_{test}$ is an imbalanced test set using a real-world corpus, which contains the common words.
In contrast, RS$_{test}$ is a balanced test set but failing to preserve the contexts.
Since WS$_{test}$ maintains the contexts, the accuracy is an important evaluation metric in WS$_{test}$ since it requires a model to predict all characters of a given word correctly.
However, WS$_{test}$ does not contain sufficient number of \emph{few} characters.
On the other hand, RS$_{test}$ is a balanced test set at the character level, so the char F1 score is a more meaningful evaluation metric compared to the accuracy.
Therefore, we measure only accuracy for WS$_{test}$ and only char F1 score for RS$_{test}$, which are collectively referred to as `\textbf{Synth}$_{test}$' in our experiments.

\begin{table*}[t!]
    \small
    \centering
    \begin{tabular}{@{}c@{}c@{}c@{}|ccccc|c@{}}
        \toprule
        \multirowcell{1}{\multicolumn{1}{c}{\textbf{Lang.}}} & 
        \multirowcell{1}{\multicolumn{1}{c}{\textbf{Test Data}}} & 
        \multirowcell{1}{\multicolumn{1}{c}{\textbf{Metric}}} & 
        Softmax & Focal & $\tau$-norm &
        PC-Sofmtax & Bal-Softmax & Ours\\
        \midrule
        \multirowcell{8}{\textit{Kr}}
         & \multirowcell{2}{\textbf{Real}} 
         & Acc & 77.43 & 77.10 & 77.59 & 77.51 & 78.37 & \textbf{81.35} \\
         && Char F1 & 0.66/0.79/\textbf{0.88} & 0.60/0.75/0.86 & 0.68/0.80/\textbf{0.88} & 0.65/0.78/0.87 & 0.62/0.76/0.87 & \textbf{0.69}/\textbf{0.81}/\textbf{0.88} \\
        \cmidrule(lr){2-9}
         & \multirowcell{2}{\textbf{Real}$_{Easy}$} 
         & Acc & 77.87 & 77.52 & 78.02 & 77.94 & 78.75 & \textbf{81.75} \\
         && Char F1 & ---/0.80/0.88 & ---/0.76/0.86 & ---/\textbf{0.81}/0.88 & ---/0.78/0.88 & ---/0.77/0.87 & ---/\textbf{0.81}/\textbf{0.89} \\
        \cmidrule(lr){2-9}
         & \multirowcell{2}{\textbf{Real}$_{Hard}$} 
         & Acc & 61.25 & 61.18 & 61.46 & 61.23 & 63.86 & \textbf{66.68} \\
         && Char F1 & 0.79/0.78/0.77 & 0.75/0.72/0.73 & 0.79/\textbf{0.79}/0.77 & 0.80/0.78/0.77 & 0.79/0.73/0.75 & \textbf{0.80}/0.75/\textbf{0.78} \\
        \cmidrule(lr){2-9}
         & \multirowcell{2}{\textbf{Synth}$_{test}$}  
         & Acc & 86.37 & 84.63 & 86.34 & 86.04 & 86.84 & \textbf{88.93} \\
         && Char F1 & 0.86/0.85/\textbf{0.83} & 0.85/0.84/0.81 & \textbf{0.87}/\textbf{0.86}/0.82 & 0.86/0.85/\textbf{0.83} & 0.86/0.85/0.82 & \textbf{0.87}/0.85/0.79 \\
        \midrule
        \multirowcell{8}{\textit{Cn}}
         & \multirowcell{2}{\textbf{Real}} 
         & Acc & 41.83 & 41.45 & 41.85 & 41.74 & 41.26 & \textbf{47.67} \\
         && Char F1 & 0.48/0.54/0.57 & 0.45/0.50/0.54 & \textbf{0.49}/\textbf{0.55}/\textbf{0.59} & 0.47/0.52/0.56 & 0.47/0.52/0.55 & 0.48/0.53/0.57 \\
        \cmidrule(lr){2-9}
         & \multirowcell{2}{\textbf{Real}$_{Easy}$} 
         & Acc & 41.72 & 41.48 & 41.76 & 41.63 & 41.14 & \textbf{48.09} \\
         && Char F1 & ---/0.55/0.58 & ---/0.52/0.55 & ---/\textbf{0.57}/\textbf{0.60} & ---/0.54/0.57 & ---/0.53/0.56 & ---/0.55/0.58 \\
        \cmidrule(lr){2-9}
         & \multirowcell{2}{\textbf{Real}$_{Hard}$} 
         & Acc & 42.71 & 41.24 & 42.57 & 42.59 & 42.25 & \textbf{44.34} \\
         && Char F1 & 0.57/\textbf{0.60}/0.55 & 0.54/0.56/0.53 & 0.57/\textbf{0.60}/\textbf{0.56} & 0.57/0.59/0.55 & \textbf{0.58}/0.59/0.55 & 0.53/0.58/0.55 \\
        \cmidrule(lr){2-9}
         & \multirowcell{2}{\textbf{Synth}$_{test}$}  
         & Acc & 83.31 & 80.06 & 83.18 & 83.02 & 82.07 & \textbf{86.22} \\
         && Char F1 & \textbf{0.83}/\textbf{0.83}/\textbf{0.78} & 0.80/0.79/0.76 & \textbf{0.83}/\textbf{0.83}/\textbf{0.78} & 0.82/0.82/\textbf{0.78} & 0.82/\textbf{0.83}/\textbf{0.78} & \textbf{0.83}/\textbf{0.83}/0.73 \\
        \bottomrule
    \end{tabular}
    \vspace{-0.25cm}
    \caption{Comparison with long-tailed recognition baselines and our method trained on CS, where we employ TRBA~\cite{baek2019STRcomparisons} for STR framework. Char F1 scores represent the scores of \emph{few} / \emph{medium} / \emph{many} characters, respectively. Our method achieves the state-of-the-art STR performance on languages including a long-tailed distribution of characters.}
    \vspace{-0.65cm}
    \label{Table:experiment-model}
\end{table*}

\noindent\textbf{Effectiveness of CAFE-Net.}
We implement four models for the experiments; (i) CNN-based STR model: TRBA~\cite{baek2019STRcomparisons} and TextAdaIN~\cite{nuriel2022textadain}, (ii) ViT-based STR model: ViTSTR+Linear and ViTSTR+Attn~\cite{atienza2021vision}.
Table~\ref{Table:experiment-ours} demonstrates that integrating CAFE-Net improves the accuracy consistently in evaluation datasets in both Korean and Chinese datasets, except for TextAdaIN+Ours on Chinese \textbf{Real}$_{Hard}$.
We want to emphasize that our method leads to a large performance improvement compared to utilizing only a long-tailed dataset (\textit{e.g.}, WS), which is widely used in the STR field.
These results demonstrate that appropriately solving the character-level long-tailed distribution can enhance overall performance for languages with a large number of characters.
Notably, our method can achieve consistent performance improvement regardless of the model architecture, demonstrating its wide applicability.

\noindent\textbf{Comparison with Baselines.}
A myriad of methods for handling long-tailed distribution datasets~\cite{lin2017focal,kang2019decoupling,hong2021disentangling,ren2020balanced} have been introduced in recent years.
Since we tackle the long-tailed distribution of characters in STR, we compare our proposed method with the existing long-tailed recognition approaches.
For the long-tailed recognition approaches, we adopt the simple techniques that are possible to apply to the STR model: (1) Softmax: the model is trained with the standard cross-entropy loss, (2) Focal loss~\cite{lin2017focal}: relatively easy classes (\textit{i.e.}, many characters) are de-emphasized, (3) $\tau$-Normalization~\cite{kang2019decoupling}: the weights of classifier are normalized with the hyper-parameter $\tau$, (4) PC Softmax~\cite{hong2021disentangling}: the logits are modified based on the label distribution during inference, (5) Balanced Softmax~\cite{ren2020balanced}: adjusting the output logits using the training label distribution.
In this experiment, we apply the baselines to TRBA model~\cite{baek2019STRcomparisons}.
The implementation details of baselines and our method are described in the supplementary.
For a fair comparison with our method, we train the TRBA~\cite{baek2019STRcomparisons} model using CS.

\begin{figure}[t!]
    \centering
    \includegraphics[width=1.0\linewidth]{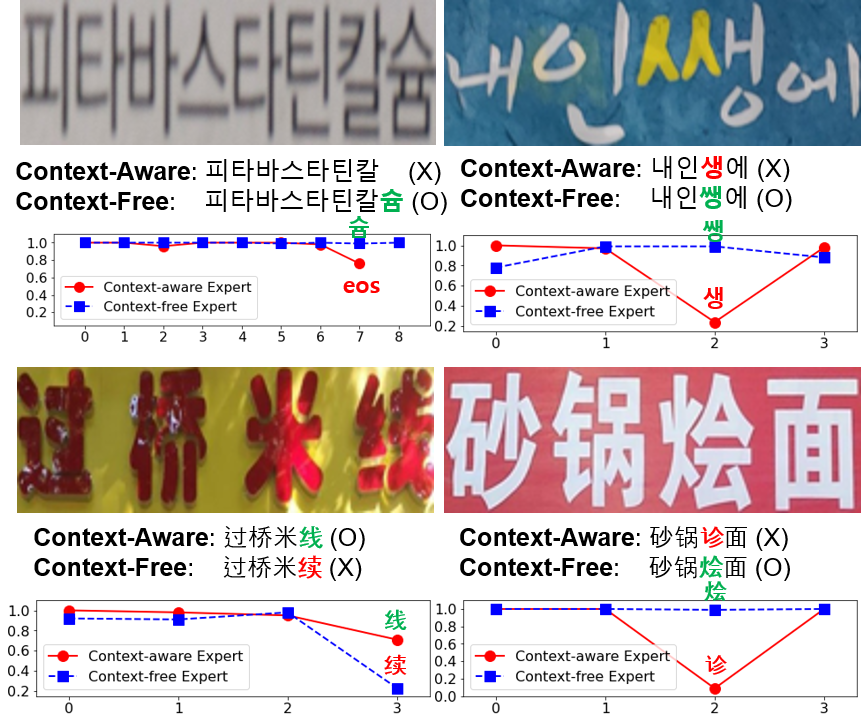}
    \vspace{-0.7cm}
    \caption{The left and right column indicates the correctly predicted samples of the context-aware and context-free expert, respectively. For each plot under each image, the x-axis and the y-axis indicate character sequence and maximum softmax probability of each character, respectively.}
    \vspace{-0.7cm}
    \label{fig:qualitative analysis}
\end{figure}

Table~\ref{Table:experiment-model} provides the summary of the performances of baselines and our method.
The results demonstrate that our method outperforms the baselines in accuracy significantly, while showing comparable performance in char F1 score.
While $\tau$-norm~\cite{wang2020decoupled} generally achieves the best char F1 score, it shows degraded performance in accuracy.
Such a result shows that the model fails to learn the contextual information, even with improved char F1 score. 
CAFE-Net, however, shows comparable char F1 score (visual representation) while achieving the best accuracy (contextual representation).
This result demonstrates the motivation of our work, which is to improve both contextual and visual representation for enhancing performance on STR with languages including numerous number of characters.

\begin{figure}[t!]
    \centering
    \includegraphics[width=0.85\linewidth]{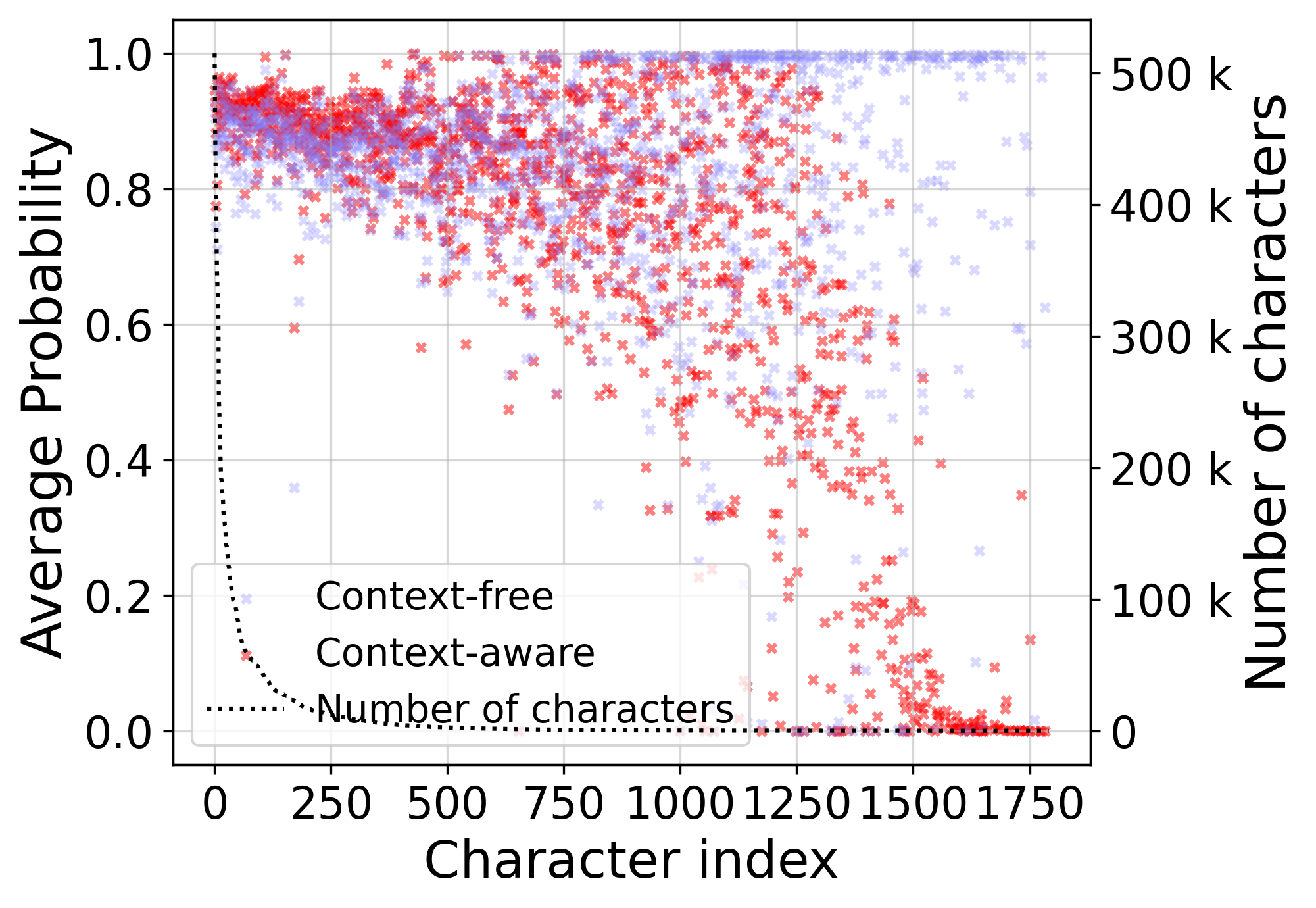}
    \vspace{-0.4cm}
    \caption{Visualization of relation between the characters and the output probability of each expert on \textbf{Real}. X-axis and y-axis indicate the characters sorted by the number of samples and the averaged probability of each character, respectively. The red and the blue dots indicate the probability of the context-aware and context-free experts, respectively.}
    \vspace{-0.6cm}
    \label{fig:confidence analysis}
\end{figure}

\noindent\textbf{Analysis on Confidence Score.}
To better comprehend why confidence ensemble has the capability to appropriately select the expert, we study the confidence score qualitatively and quantitatively.
Fig.~\ref{fig:qualitative analysis} shows the prediction and the maximum softmax probability of each expert on several samples.
Since the context-free expert focuses on the visual representation, it mispredicts confusing characters (Fig.~\ref{fig:qualitative analysis} left column).
In contrast, we observe that context-aware expert incorrectly predicts \emph{few} characters as \emph{many} characters by resorting to the context when making predictions (Fig.~\ref{fig:qualitative analysis} right column).
We observe that each expert outputs low maximum softmax probability with confusing samples (\textit{e.g.}, visually confusing character for context-free expert, and \textit{few} characters for context-aware expert).
Our confidence ensemble enables to filter out such low-confident prediction of one expert and select the high-confident prediction of the other expert, improving the STR performance overall.

Fig.~\ref{fig:confidence analysis} visualizes the averaged prediction probability at the ground truth character. 
We observe that the context-aware expert (red) achieves higher prediction probability with \textit{many} classes than the context-free expert (blue). 
On the other hand, the context-free expert shows higher prediction probability with large margin on the \textit{few} characters compared to the ones of context-aware expert. 
Such a visualization demonstrates that confidence ensemble enables the two experts to compensate the limitation of each other.

\begin{figure}[t!]
    \centering
    \includegraphics[width=1.0\linewidth]{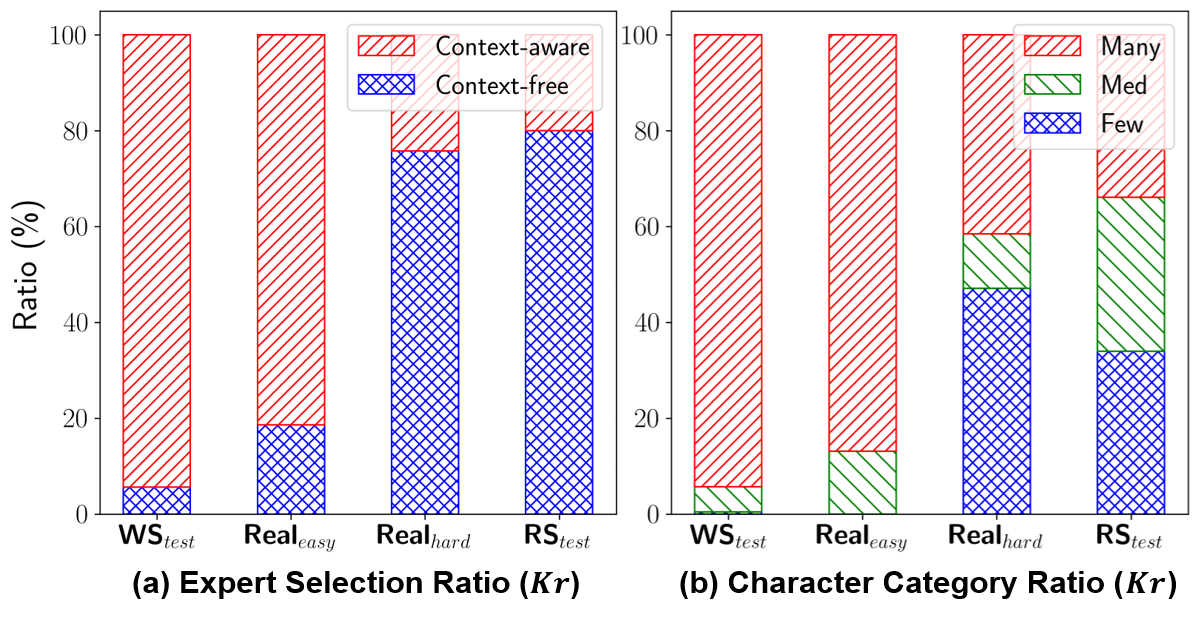}
    \vspace{-0.8cm}
    \caption{(a) We observe that CAFE-Net selects more predictions from context-aware branch for WS$_{test}$ and \textbf{Real}$_{Easy}$, while selecting more predictions from context-free branch for \textbf{Real}$_{Hard}$ and RS$_{test}$. CA and CF indicates the context-aware branch and the context-free branch, respectively. (b) We show the proportion of \emph{many}, \emph{medium}, and \emph{few} characters in each test set.}
    \vspace{-0.6cm}
    \label{fig:ratio analysis}
\end{figure}

\noindent\textbf{Expert Selection Ratio.}
We analyze the relation between the proportion of the samples allocated to each expert and the character category ratio in the test sets.
Interestingly, we discover that the ratio of predictions selected by the context-aware expert in dataset is proportional to the ratio of \emph{many} characters in a dataset as shown in Fig.~\ref{fig:ratio analysis}.
In summary, these results indicate that the context-free expert tends to predict the instances containing \emph{few} or \emph{medium} characters, whereas the context-aware expert predicts the rest of the instances including only \emph{many} characters more frequently.
We also report the accuracy of each expert in the supplementary.

\noindent\textbf{Effectiveness of Confidence Ensemble.}
In Table~\ref{Table:experiment-ensemble}, we show that careful consideration regarding how to ensemble two different experts is important.
We observe that utilizing our method, a word-level confidence ensemble, outperforms the character-level confidence ensemble, which aggregates the outputs at the character level using the maximum softmax probability.
The main reason is that the word-level ensemble performs more robustly than the character-level ensemble when misalignment happens between the predicted words by two experts.
As shown, while ensemble may be a straightforward and widely used approach, considering such a property for scene text recognition is important. 
We want to emphasize that our method well reflects such characteristic and improves STR performance.

\begin{figure}[t!]
    \centering
    \includegraphics[width=1.0\linewidth]{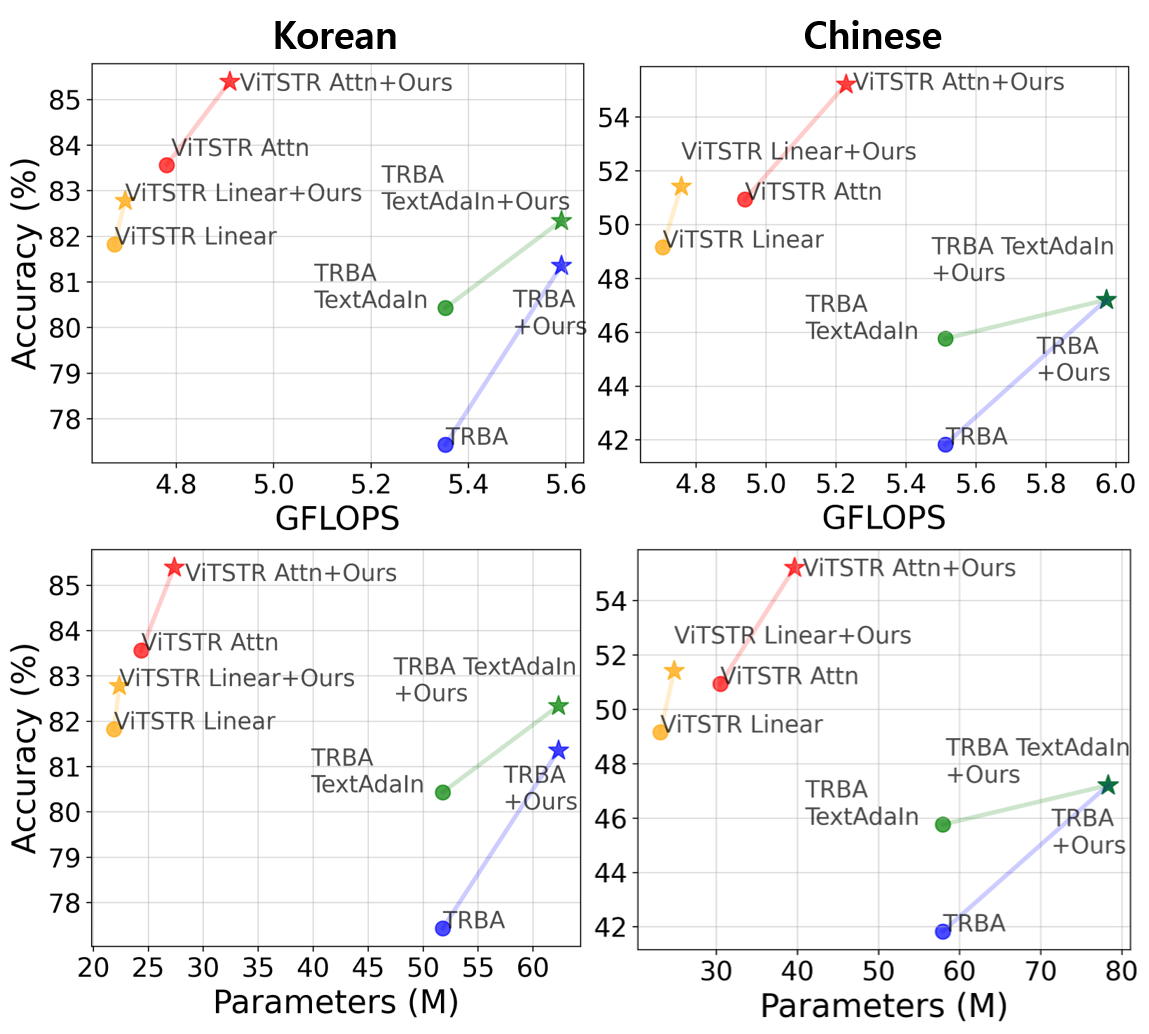}
    \vspace{-0.7cm}
    \caption{Comparisons on accuracy and computational costs. We use \textbf{Real} dataset for the analysis.}    
    \vspace{-0.3cm}
    \label{fig:cost}
\end{figure}

\noindent\textbf{Computational Cost.}
Fig.~\ref{fig:cost} summarizes the accuracy on \textbf{Real} dataset and the computational costs (\textit{e.g.}, flops and the number of parameters).
While applying our method consistently improves performance regardless of the model architectures, we observe that our method requires a negligible amount of additional computational costs.
The main reason is that we only require an additional classifier, which occupies a negligible amount of weight parameters. 
For example, about 1\% flops and 3$\sim$7\% parameters increase when applying our method to ViTSTR~\cite{atienza2021vision}.

\begin{table}[t!]
    \small
    \centering
    \begin{tabular}{cl|p{0.07\textwidth}<{\centering}p{0.07\textwidth}<{\centering}p{0.07\textwidth}<{\centering}}
        \toprule
        \textbf{Lang.} & \multicolumn{1}{c}{\textbf{Method}} & \textbf{Real} & \textbf{Real$_{Easy}$} & \textbf{Real$_{Hard}$} \\
        \midrule
        \multirowcell{2}{\textit{Kr}}
        & Char-level & 80.62 & 81.20 & 59.07 \\
        & Word-level & \textbf{81.35} & \textbf{81.75} & \textbf{66.68} \\
        \midrule
        \multirowcell{2}{\textit{Cn}}
        & Char-level & 45.36 & 47.42 & 29.14 \\
        & Word-level & \textbf{47.67} & \textbf{48.09} & \textbf{44.34} \\
        \bottomrule
    \end{tabular}
    \vspace{-0.3cm}
    \caption{Ablation study on ensemble technique using \textbf{Real} Korean and Chinese datasets. TRBA model~\cite{baek2019STRcomparisons} is utilized.}
    \vspace{-0.5cm}
    \label{Table:experiment-ensemble}
\end{table}

\section{Conclusions}

This paper investigates character-level long-tailed distribution in STR, which has been overlooked in STR previously.
Our empirical analysis indicates that improving both contextual and visual representation is crucial for improving STR on languages including characters with long-tailed distribution. 
Based on the finding, we propose a Context-Aware and Free Experts Network (CAFE-Net), which trains two different experts to focus on learning contextual information and visual representation, respectively.
To aggregate two different experts, we propose the confidence ensemble to improve STR performance on all \emph{many}, \emph{medium}, and \emph{few} characters.
Extensive experiments show that we achieve the state-of-the-art performance with languages showing the long-tailed distributions at the character level.
We believe that our work inspires the future researchers to improve STR on languages with numerous characters, which is relatively under-explored compared to STR on English.

{\small
\bibliographystyle{ieee_fullname}
\bibliography{reference}
}

\twocolumn[
\begin{center}
    \vspace*{1.1cm}
    \Large{\bf{Supplementary Material}}
    \vspace*{1.7cm}
\end{center}]

\section*{A. Additional Details}

In this section, we describe the details of the experiments.
First, we describe the additional details for the experiments in Section~3 (Section~A.1).
Moreover, we explain the details of the character-level (char) F1 score (Section~A.2).
Then, we describe the pre-processing details for the datasets we utilized (Section~A.3).

\subsection*{A.1. Experimental Setup in Section~3}

For the experiments in Section~3, we utilize a widely-used model architecture in STR, denoted as TRBA~\cite{baek2019STRcomparisons}.
Note that we use the same model architecture for Section~3 experiments for a fair comparison.
According to the previous work~\cite{baek2019STRcomparisons}, the four stages derived from the STR models are as follows:
\begin{itemize}
    \item \textbf{Transformation.} The thin-plate spline (TPS) transformation, a variant of a spatial transformation network~\cite{jaderberg2015spatial}, normalizes the perspective or curved text image into a horizontal text image.
    \item \textbf{Feature Extraction.} A convolutional neural network maps the image to the visual feature representation. In our experiments, we adopt ResNet~\cite{he2016deep} for the feature extractor, which was widely used in previous studies~\cite{baek2019STRcomparisons}.
    \item \textbf{Sequence Modeling.} After the feature extraction stage, we obtain contextual features using a sequence model such as BiLSTM to improve the representation.
    \item \textbf{Prediction.} We employ the attention-based decoder to predict the sequence of characters since it shows superior performance compared to the CTC decoder~\cite{graves2006connectionist}.
\end{itemize}

As described in the main paper, we evaluate the models by individually training them with WikiSynth (WS), RandomSynth (RS), and CombinedSynth (CS) using Korean.
The objective function for training is equal to the previous work~\cite{baek2019STRcomparisons} as follows:
\begin{equation}
    \mathcal{L} = -\frac{1}{T} \sum^{T}_{t=1} \log p(y^t|h^t),
\end{equation}
where $y^t$ is the $t$-th ground truth character.
To train the models, we utilize Adadelta~\cite{zeiler2012adadelta} optimizer with a learning rate of 1.0.
We train the models for 200,000 iterations.
The batch size is set to 192.
Moreover, we conduct all experiments using NVIDIA A100 GPU.

\subsection*{A.2. Character-Level F1 Score}

\begin{figure}[t!]
    \centering
    \includegraphics[width=1.0\linewidth]{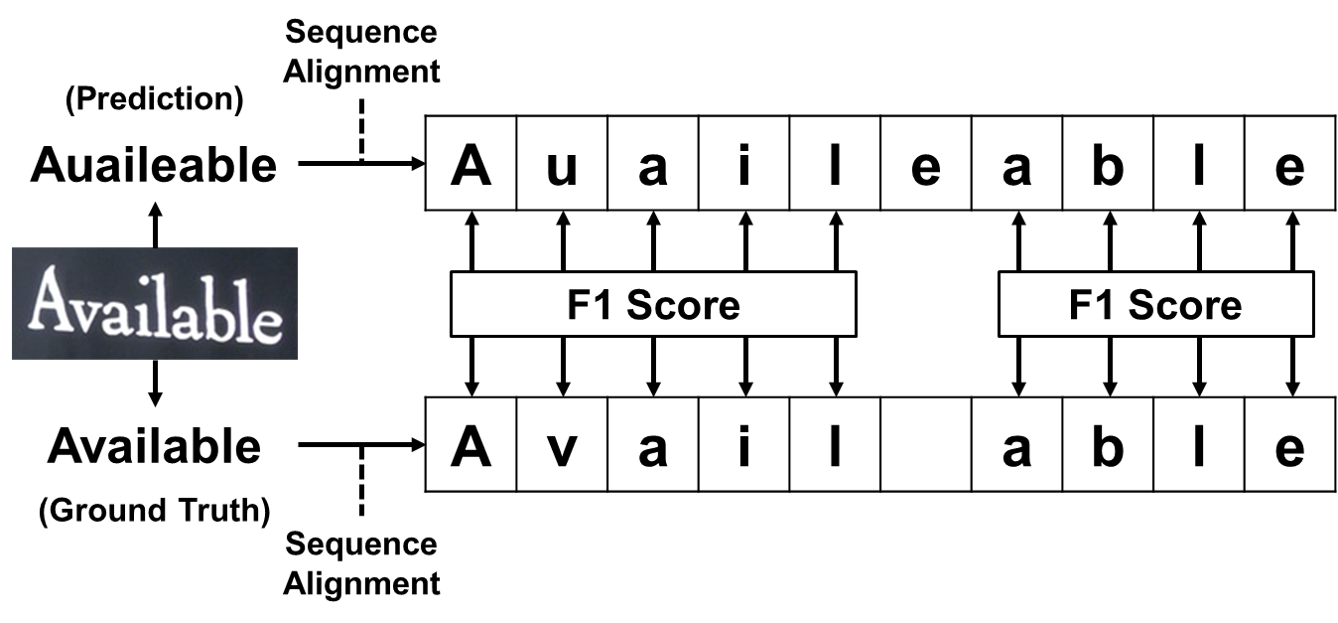}
    \caption{Diagram for computation of char F1 score.}
    \vspace{-0.3cm}
    \label{supp-fig:char f1 score}
\end{figure}

Fig.~\ref{supp-fig:char f1 score} shows the process of computation of character-level (char) F1 score.
We compute the char F1 score in the following process.
First, we perform the sequence alignment of ground truth and predicted character using Hirschberg's algorithm~\cite{hirschberg1975linear}.
Hirschberg's algorithm is one of the techniques to find the alignment path.
After aligning the predicted and the ground truth words, we compute F1 score per character as follows:
\begin{equation*}
    \text{F1 score} = 2 \times \frac{\text{recall} \times \text{precision}}{\text{recall} + \text{precision}}.
\end{equation*}
Since the F1 score is more suitable than the accuracy for the imbalanced number of data samples~\cite{haibo2013imbalanced}, we compute the F1 score instead of accuracy per character for evaluating STR performance on individual characters.
Finally, we average these scores per character.
Then, as described in the main paper, we categorize the characters into three groups: 1) \emph{many} (\textit{i.e.,} $n_i \geq 1500$), 2) \emph{medium} (\textit{i.e.,} $n_i<1500$ and $n_i \geq 100$), and 3) \emph{few} (\textit{i.e.,} $n_i<100$).

\begin{figure}[t!]
    \centering
    \includegraphics[width=1.0\linewidth]{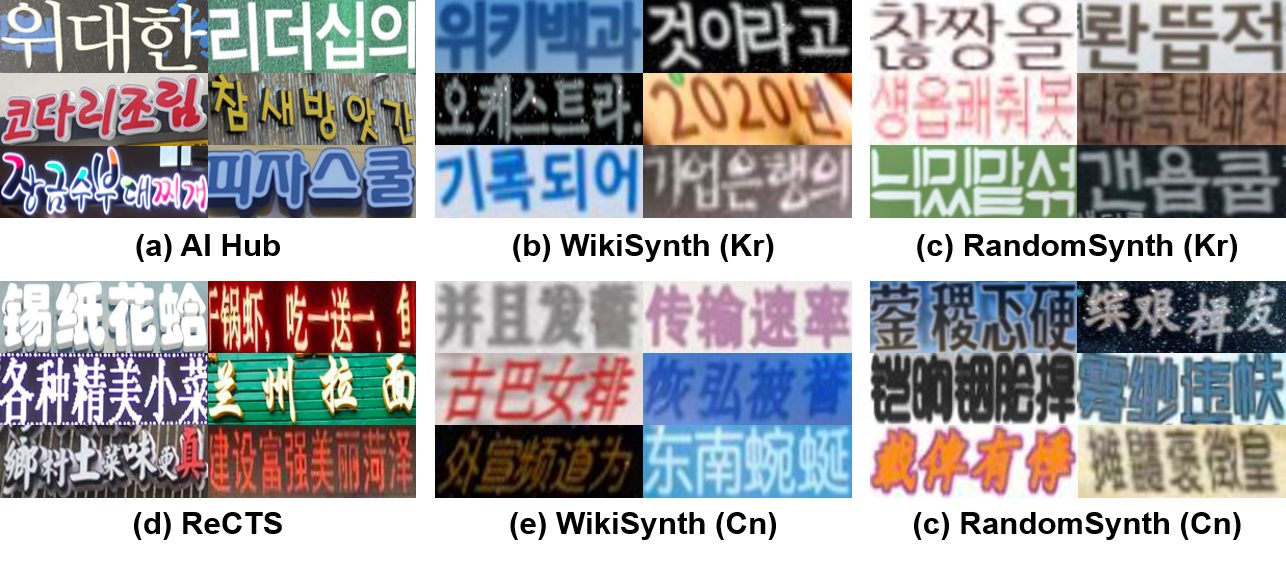}
    \vspace{-0.6cm}
    \caption{Examples of Korean and Chinese datasets. AI Hub and ReCTS datasets are real-world test sets termed as `\textbf{Real}'. WikiSynth (WS) and RandomSynth (RS) are synthetic datasets, which show different character-level distributions.}
    \vspace{-0.5cm}
    \label{supp-fig:dataset}
\end{figure}

\subsection*{A.3. Pre-processing Dataset}

We describe the details of the \textbf{Real} datasets utilized in our experiments.
AI Hub, a Korean real STR dataset, includes 151,105 cropped images of street signs, traffic signs, and brands.
ReCTS, a Chinese real STR dataset, contains 77,709 cropped images of Chinese signboards in the street view with diverse backgrounds and fonts.
Furthermore, the images in AI Hub and ReCTS are pre-processed using the following procedures to be used for the text recognition task.
First, word boxes are cropped, and images including characters beyond the predefined set are filtered out.
In addition, to filter vertical images, we excluded images with three or more characters and heights greater than the widths.
Fig.~\ref{supp-fig:dataset} shows the sample images in the datasets utilized in the experiments.

\section*{B. Implementation Details of CAFE-Net}

Since the proposed method is easily applicable to various STR models, we introduce two versions of our method: (i) CNN-based CAFE-Net, which includes CNN-based feature extractor and (ii) ViT-based CAFE-Net, which contains a vision transformer encoder~\cite{vaswani2017attention,dosovitskiy2020vit,atienza2021vision}.
Since these two types of backbones are widely used in STR field, we describe our method based on these two architectures.
Moreover, our code will be released upon paper acceptance.

\begin{figure}[t!]
    \centering
    \includegraphics[width=1.0\linewidth]{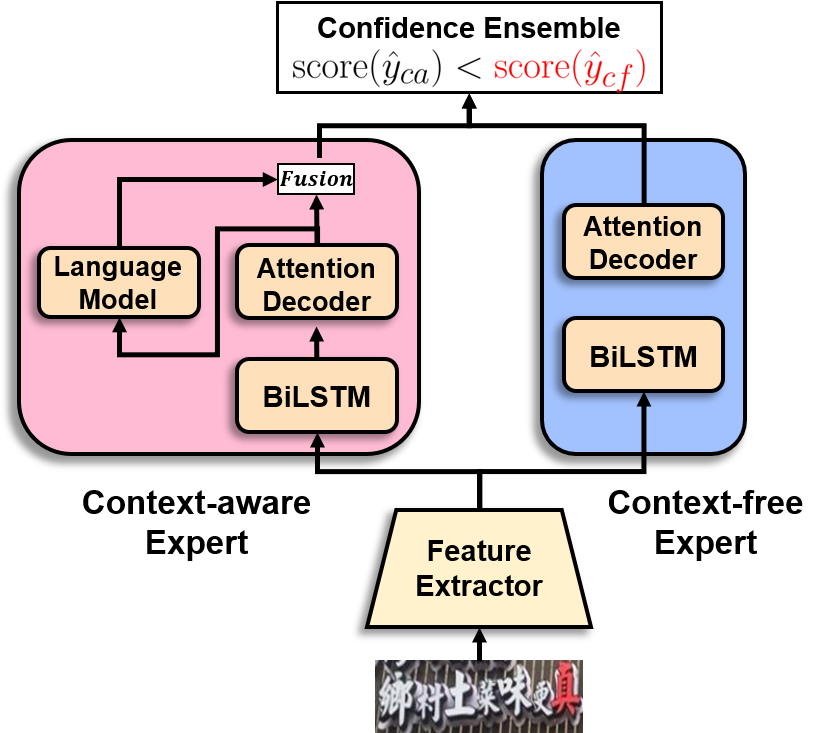}
    \vspace{-0.5cm}
    \caption{Overview of CNN-based CAFE-Net.}
    \vspace{-0.4cm}
    \label{supp-fig:model-cnn}
\end{figure}

\noindent\textbf{CNN-based CAFE-Net.}
Fig.~\ref{supp-fig:model-cnn} shows the overview of CNN-based CAFE-Net such as TRBA + Ours and TextAdaIN + Ours.
Referring to the previous STR literature~\cite{baek2019STRcomparisons}, CNN-based STR models generally consist of CNN-based feature extractor (\textit{e.g.}, ResNet~\cite{he2016deep}), the sequence modeling (\textit{e.g.}, BiLSTM) and prediction layer (\textit{e.g}, CTC~\cite{graves2006connectionist} and Attention~\cite{cheng2017focusing} decoders).
In our framework, two experts share the feature extractor.
Sharing weights largely reduces the computational complexity in the inference phase.
Each expert in CNN-based CAFE-Net consists of the sequence modeling and prediction layers.

The training images $x_{ca}$ and $x_{cf}$ sampled for the context-aware and the context-free experts are fed into the feature extractor to acquire feature representations $f_{ca}$ and $f_{cf}$, respectively.
Given the feature representation $f_{cf}$, a context-free expert produces the output feature $h_{cf}=\{ h^{(1)}_{cf}, \dots, h^{(T)}_{cf} \}$ of the corresponding words $\hat{y}_{cf}=\{ \hat{y}^{(1)}_{cf}, \dots, \hat{y}^{(T)}_{cf} \}$.
Here, $T$ denotes the maximum length of the word.
The context-free expert is trained using only RS to improve the performance on \emph{few} characters.
Due to the balanced number of characters, the context-free expert correctly predicts \emph{few} characters more than the context-aware expert.
As mentioned in the main paper, this is mainly due to the fact that random sequences of characters enforces the model to focus on the visual representation rather than the contextual representation.

Different from the context-free expert, the context-aware expert is trained with WS to focus on learning the contextual information, which is essential to predict the whole words accurately.
Moreover, following a recent context-aware STR method~\cite{yu2020towards} that is one of CNN-based STR models, we leverage an external language model to capture semantic information to assist STR.
We simply add a transformer as an external language model.
Specifically, with the feature representations $f_{ca}$ and $f_{cf}$, the context-aware expert produces the output feature.
Then, an external language model refines the output of the context-aware expert.
Finally, the outputs of the context-aware expert and the language model are fused to produce the final output feature.
In summary, the context-aware expert with the external language model produces the final output feature $h_{ca}=\{ h^{(1)}_{ca}, \dots, h^T_{ca} \}$ of the corresponding words $\hat{y}_{ca}=\{ \hat{y}^{(1)}_{ca}, \dots, \hat{y}^{(T)}_{ca} \}$.

A context-free expert is trained using a cross-entropy loss, as described in the paper.
In contrast, a context-aware expert is trained using three loss functions, following the objective function of semantic reasoning network~\cite{yu2020towards}.
Specifically, the cross-entropy loss is applied to the outputs of the attention decoder, the language model, and the final output $\hat{y}_{ca}$, respectively.

TextAdaIN + Ours model is implemented by applying TextAdaIN to the feature extractor (\textit{i.e.}, ResNet).
Furthermore, Adadelta~\cite{zeiler2012adadelta} is utilized for training our model with the learning rate of 2.0.
We train the model for 200,000 iterations.
The batch size is set to 192 for each expert.

\noindent\textbf{ViT-based CAFE-Net.}
Based on a ViTSTR~\cite{atienza2021vision} that is a vision transformer for STR task, we design a ViT-based CAFE-Net.
Different from CNN-based CAFE-Net, the input images $x_{ca}$ and $x_{cf}$ are reshaped into a sequence of flattened 2D $P \times P$ patches, where $P$ denotes the patch size.
The positional encoding is added to the flattened input patches.
Moreover, a learnable patch embedding vector $x_{class}$ is appended to the input for the transformer encoder.
Formally, the input of the encoder is as follows:
\begin{equation}
    z_0 = [x_{class}; x^{(1)}\textbf{E}, x^{(2)}\textbf{E}, \cdots, x^{(N)}\textbf{E}] + \textbf{E}_{pos},
\end{equation}
where $\textbf{E}$ and $\textbf{E}_{pos}$ denote a linear projection layer and a positional encoding, respectively.
$N$ indicates the number of input patches.
Given $z^{0}$, the transformer encoder produces the feature maps $z^{K}$, which are the outputs of $K$-th transformer block.
Here, $K$ denotes the number of blocks in the transformer encoder.

In the ViT-based CAFE-Net, each expert consists of the module for reshaping the output of the transformer encoder and the prediction part, such as an attention decoder for predicting the words.
To make $z^K$ to $h=\{ h^{(1)}, \dots, h^T \}$, we utilize a module, which consists of the attention layer such as a parallel visual attention module in previous STR work~\cite{yu2020towards}.
Finally, the prediction parts in the context-free and context-aware experts predict $\hat{y}_{cf}$ and $\hat{y}_{ca}$ with $h_{cf}$ and $h_{ca}$.
Unlike CNN-based CAFE-Net, we do not employ an external language model.
Adam~\cite{kingma2014adam} optimizer is employed for training the model with the learning rate of 5e-4.
To ensure the stability of model training, we utilize a cosine annealing scheduler to adjust the learning rate.
The patch size $P$ is set to 4 and the number of transformer block $K$ is set to 12. 
We train the model for 200,000 iterations.
The batch size is set to 192 for each expert.

In our experiments, we implement two types of ViT-based CAFE-Net: (i) ViTSTR + Linear and (ii) ViTSTR + Attn.
First, ViTSTR + Linear consists of the vision transformer encoder and linear decoder, which contains a fully-conntected layer following the vision transformer encoder.
On the other hand, ViTSTR + Attn utilizes the attention decoder instead of the linear layer.
The primary distinction between two models is that ViTSTR + Linear model predicts the words in parallel, while ViTSTR + Attn produces the characters autoregressively.

\begin{figure}[t!]
    \centering
    \includegraphics[width=1.0\linewidth]{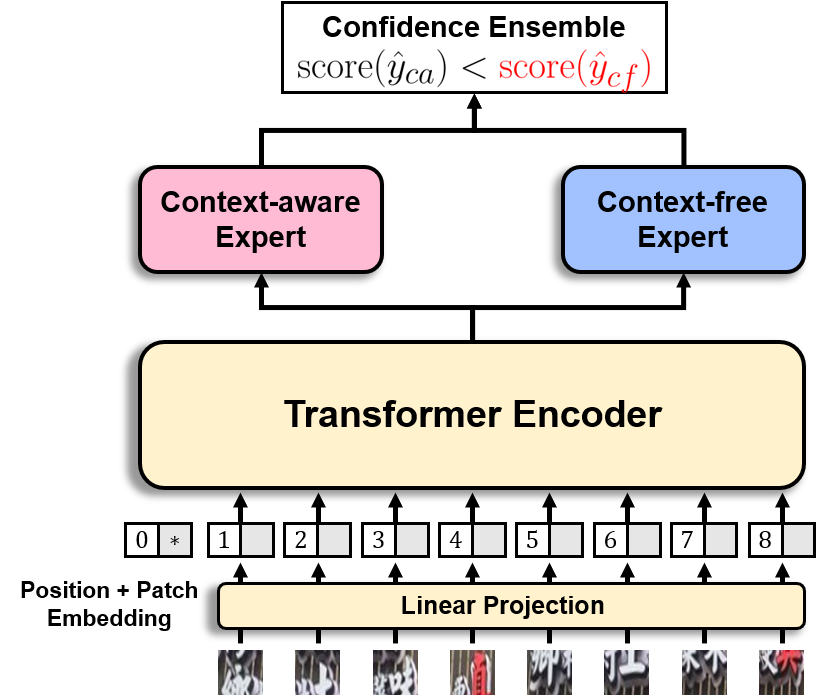}
    \vspace{-0.5cm}
    \caption{Overview of ViT-based CAFE-Net.}
    \vspace{-0.5cm}
    \label{supp-fig:model-vit}
\end{figure}

\section*{C. Baselines}

\noindent\textbf{Focal Loss.}
The objective of focal loss~\cite{lin2017focal} is to emphasize relatively hard examples and enforce a model to learn the minority classes.
Formally, the focal loss adds a modulating factor $(1-p)^{\gamma}$ to the standard cross entropy criterion, where $\gamma$ is a hyper-parameter to adjust the rate of de-emphasizing the majority classes. 
The focal loss is described as:
\begin{equation*}
  \mathcal{L}_{\text{focal}} = -(1-p)^{\gamma}\log(p),
\end{equation*}
where $\gamma$ is set to 1 in our experiments.

\noindent\textbf{$\tau$-Normalization.}
Inspired by the observation that the norms of the weights $\| w_i \|$ are correlated with the frequency of the classes, $\tau$-normalization~\cite{kang2019decoupling} is proposed to adjust the classifier weight norms directly.
Formally, $\tau$-normalization is described as:
\begin{equation*}
  \Tilde{w}_i = \frac{w_i}{\| w_i \|^\tau},
\end{equation*}
where $\tau$ denotes the hyper-parameter for controlling the temperature of the normalization.
We set the hyper-parameter $\tau$ as 1 following the original paper.

\noindent\textbf{PC Softmax.}
To handle the data distribution shift, post-compensation softmax (PC softmax)~\cite{hong2021disentangling} modifies the model logits during inference.
Formally, as follows:
\begin{equation*}
    p(y|x) = \frac{e^{(f_{\theta}(x)[y] - \log{p_s(y)} + \log{p_t(y)})}}{\Sigma_c {e^{(f_{\theta}(x)[c] - \log{p_s(c)} + \log{p_t(c)})}}},
\end{equation*}
where $p_s(y)$ and $p_t(y)$ denotes the source and target data distribution, respectively. 
$f_{\theta}(x)[y]$ is the logit of class $y$ from the softmax.
In the inference phase, we utilize the modified logit $p(y|x)$ to match the target label distribution $p_t(y)$ (\textit{i.e.,} we use a balanced distribution as $p_t(y)$ in our experiments).

\begin{table*}[t!]
    \small
    \centering
    \begin{tabular}{clccccccc@{}c}
    \toprule
        \multirowcell{3}{\multicolumn{1}{c}{\textbf{Data}}} & \multirowcell{3}{\multicolumn{1}{c}{\textbf{Method}}} & \multicolumn{6}{c}{\multirowcell{1}{\textbf{Real}}} & \textbf{Synth} & \\
        \cmidrule(lr){3-8} \cmidrule(lr){9-10} 
        && \multirowcell{2}{\textbf{IIIT} \\ 3000} & \multirowcell{2}{\textbf{SVT} \\ 647} & \multirowcell{2}{\textbf{IC13} \\ 1015} & \multirowcell{2}{\textbf{IC15} \\ 2077} & \multirowcell{2}{\textbf{SVTP} \\ 645} & \multirowcell{2}{\textbf{CT} \\ 288} & \multirowcell{2}{\textbf{WS$_{test}$} \\ 10000} \\ \\
        \midrule
        \multicolumn{3}{l}{\textbf{Case-insensitive (Reported Results)}} \\
        \midrule

        MJ+ST & TRBA \cite{baek2019STRcomparisons} & 87.9 & 87.5 & 92.3 & 71.8 & 79.2 & 74.0 & - \\
        MJ+ST & DAN \cite{wang2020decoupled} & 94.3 & 89.2 & 93.9 & 74.5 & 80.0 & 84.4 & - \\
        MJ+ST & RobustScanner \cite{yue2020robustscanner} & 95.3 & 88.1 & \textbf{94.8} & \textbf{77.1} & 79.5 & \textbf{90.3} & - \\
        MJ+ST & SRN \cite{yu2020towards} & 94.8 & 91.5 & - & - & 85.1 & 87.8 & - \\
        MJ+ST & ABINet \cite{fang2021read} & \textbf{96.2} & \textbf{93.5} & - & - & \textbf{89.3} & 89.2 & - \\
        MJ+ST & ViTSTR-B \cite{atienza2021vision} & 88.4 & 87.7 & 92.4 & 72.6 & 81.8 & 81.3 & - \\
        \midrule
        \multicolumn{3}{l}{\textbf{Case-insensitive}} \\
        \midrule
        MJ+WS & ViTSTR+Linear & 93.27 & 88.10 & 93.40 & 77.18 & 79.69 & 81.94 & 92.31 \\
        MJ+CS & ViTSTR+Linear & 93.47 & 87.33 & 92.22 & 76.31 & 79.07 & 84.38 & 91.87 \\
        \rowcolor{gray!10} MJ+CS & ViTSTR+Linear+Ours & 93.40 & 88.10 & 93.89 & 77.95 & 80.00 & 82.64 & 93.65 \\
        MJ+WS & ViTSTR+Attn & 94.70 & \textbf{90.11} & \textbf{94.48} & 78.86 & \textbf{83.88} & 85.42 & 94.23 \\
        MJ+CS & ViTSTR+Attn & 94.73 & 88.72 & 94.38 & 79.73 & 83.41 & \textbf{88.54} & 93.50 \\
        \rowcolor{gray!10} MJ+CS & ViTSTR+Attn+Ours & \textbf{95.03} & 88.87 & 93.60 & \textbf{80.84} & 83.57 & 86.46 & \textbf{94.98} \\
        \midrule
        \multicolumn{3}{l}{\textbf{Case-sensitive}} \\
        \midrule
        MJ+WS & ViTSTR+Linear & 88.60 & 81.92 & - & 70.25 & 69.77 & 75.69 & 90.66 \\
        MJ+CS & ViTSTR+Linear & 88.97 & 81.76 & - & 70.15 & 70.54 & 78.82 & 90.09 \\
        \rowcolor{gray!10} MJ+CS & ViTSTR+Linear+Ours & 90.13 & 83.15 & - & 72.41 & 74.26 & 76.74 & 92.26 \\
        MJ+WS & ViTSTR+Attn & 91.00 & 83.93 & - & 73.33 & 75.50 & 78.47 & 92.80 \\
        MJ+CS & ViTSTR+Attn & 91.57 & 82.54 & - & 74.00 & 75.97 & \textbf{81.25} & 91.90 \\
        \rowcolor{gray!10} MJ+CS & ViTSTR+Attn+Ours & \textbf{91.77} & \textbf{84.70} & - & \textbf{75.83} & \textbf{77.83} & 79.86 & \textbf{93.66} \\
        \bottomrule
    \end{tabular}
    \vspace{-0.3cm}
    \caption{Accuracy on English datasets. We describe the number of test samples in each dataset in 3rd row. }
    \vspace{-0.4cm}
    \label{Table:experiment-english-acc}
\end{table*}

\noindent\textbf{Balanced Softmax.}
To re-balance the data distribution, balanced softmax~\cite{ren2020balanced} accommodates the label distribution shifts between the training and test datasets.
Formally, the balanced softmax is described as:
\begin{equation*}
  \mathcal{L}_{\text{bal}} = -\log(p + \log(\pi)),
\end{equation*}
where $\pi$ denotes the frequency of the training classes.
Different from PC softmax, the balanced softmax considers the long-tailed distribution during training the model.

\section*{D. Experiments on English}

\begin{figure}[t!]
    \centering
    \includegraphics[width=0.75\linewidth]{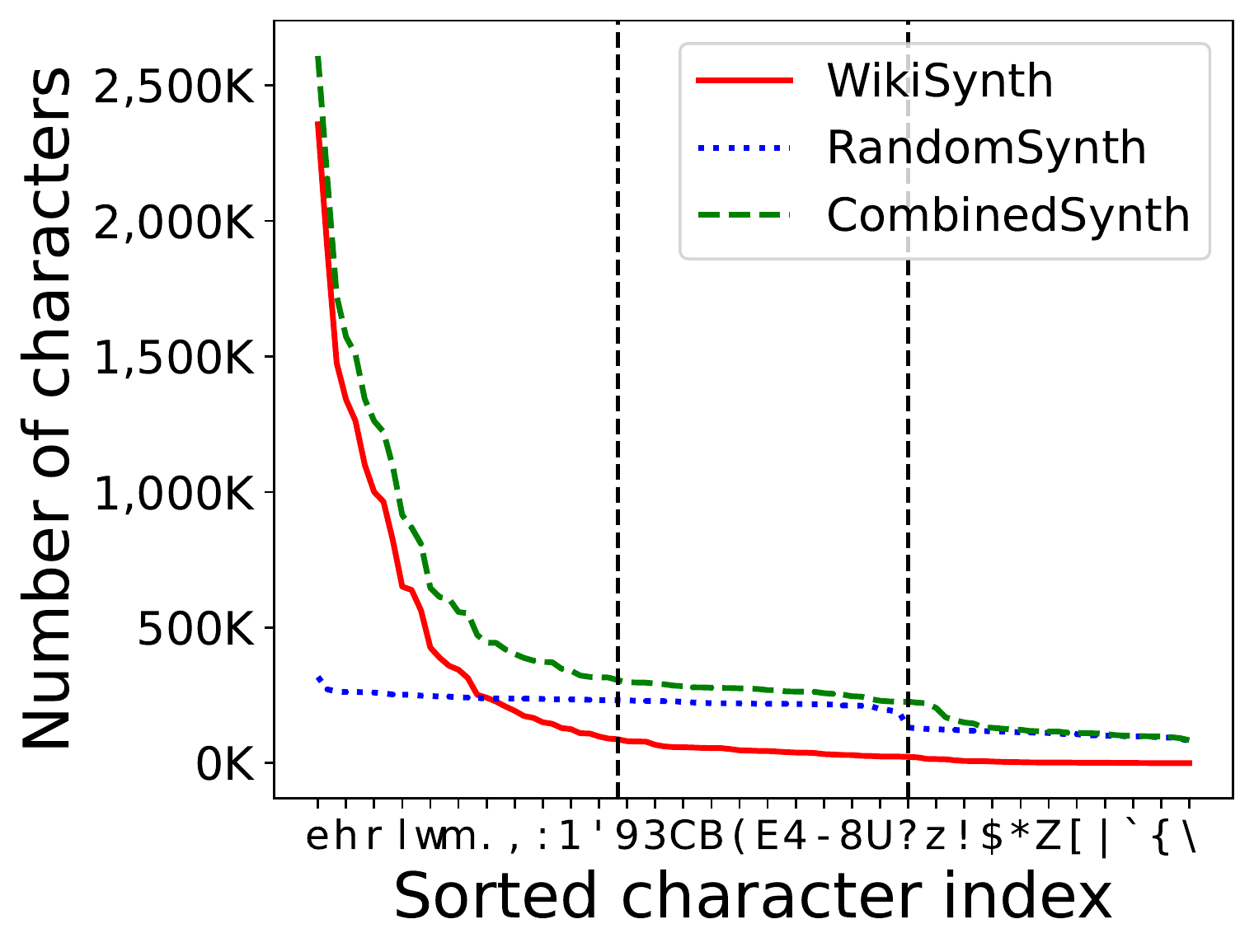}
    \vspace{-0.3cm}
    \caption{Character-level distribution of WikiSynth (WS).}
    \vspace{-0.4cm}
    \label{supp-fig:distribution-eng}
\end{figure}

To further validate the effectiveness of CAFE-Net, we conduct additional experiments on English datasets, which are widely utilized in STR fields.
To train our method, we construct the synthetic datasets (\textit{e.g.}, WS, RS, and CS) using `newsgroup' corpus, which is also utilized to synthesize the original SynthText dataset~\cite{Gupta16}.
Fig.~\ref{supp-fig:distribution-eng} shows the character-level distributions of WS, RS, and CS.
Previous STR methods~\cite{baek2019STRcomparisons,wang2020decoupled,yue2020robustscanner,yu2020towards,fang2021read,atienza2021vision} generally employ SynthText (ST) and MJSynth (MJ)~\cite{Jaderberg14c} for training the models.
In order to conduct experiments in a similar setting, we also use MJSynth for training the models.
For real test sets, we employ IIIT5k-Words (IIIT)~\cite{mishra2012scene}, Street View Text (SVT)~\cite{wang2011end}, ICDAR2013 (IC13)~\cite{karatzas2013icdar}, ICDAR2015 (IC15)~\cite{karatzas2015icdar}, SVT Perspective (SVTP)~\cite{phan2013recognizing}, and CUTE80 (CT)~\cite{risnumawan2014robust}, which are widely used in previous work.
Also, we utilize the synthetic test set, WS$_{test}$, for measuring the accuracy.

However, previous evaluations on English datasets were generally conducted using only lowercase letters and numbers (\textit{i.e.}, case-insensitive setting) without distinguishing between the small and capital letters.
In addition to the case-insensitive setting, we assess the performance in a more practically useful setting, which includes uppercase letters and symbols (\textit{i.e.}, case-sensitive setting).
The character set consists of 94 characters, including uppercase letters, lowercase letters, numbers, and symbols in the case-sensitive setting.
We take the annotations for the case-sensitive setting from the previous work~\cite{long2020unreal}.
However, in the case of IC13 dataset, there are no annotations consisting of both uppercase and lowercase letters, hence we did not evaluate the models on IC13 under the case-sensitive setting.
Moreover, we compare the reported results of the existing baselines, such as TRBA~\cite{baek2019STRcomparisons}, DAN~\cite{wang2020decoupled}, RobustScanner~\cite{yue2020robustscanner}, SRN~\cite{yu2020towards}, ABINet~\cite{fang2021read}, and ViTSTR-B~\cite{atienza2021vision}, in the case-insensitive setting.

\begin{table*}[t!]
    \small
    \centering
    \begin{tabular}{@{}ccc|ccc|c}
        \toprule
        \multirowcell{1}{\multicolumn{1}{c}{\textbf{Lang.}}} & 
        \multirowcell{1}{\multicolumn{1}{c}{\textbf{Expert}}} & 
        \multirowcell{1}{\multicolumn{1}{c}{\textbf{Metric}}} & 
        \textbf{Real} & \textbf{Real$_{Easy}$} & \textbf{Real$_{Hard}$} & \textbf{Synth$_{test}$} \\
        \midrule
        \multirowcell{7}{\textit{Kr}}
         & \multirowcell{2}{Confidence Ensemble} 
         & Acc & \textbf{81.35} & \textbf{81.75} & \textbf{66.68} & 88.93 \\
         && Char F1 & \textbf{0.69}/0.81/\textbf{0.88} & --/0.81/\textbf{0.89} & \textbf{0.80}/\textbf{0.75}/0.78 & \textbf{0.87}/\textbf{0.85}/0.79 \\
        \cmidrule(lr){2-7}
         & \multirowcell{2}{Context-Aware Expert} 
         & Acc & 80.45 & 81.14 & 54.66 & \textbf{89.27} \\
         && Char F1 & 0.51/\textbf{0.81}/0.88 & --/\textbf{0.83}/\textbf{0.89} & 0.52/0.69/0.76 & 0.37/0.72/0.71 \\
        \cmidrule(lr){2-7}
         & \multirowcell{2}{Context-Free Expert} 
         & Acc & 78.01 & 78.32 & 66.57 & 76.76 \\
         && Char F1 & 0.52/0.69/0.86 & --/0.70/0.87 & 0.80/0.74/\textbf{0.80} & 0.85/0.84/\textbf{0.85} \\
        \midrule
        \multirowcell{7}{\textit{Cn}}
         & \multirowcell{2}{Confidence Ensemble} 
         & Acc & \textbf{47.67} & \textbf{48.09} & 44.34 & \textbf{86.22} \\
         && Char F1 & \textbf{0.48}/\textbf{0.53}/0.57 & --/\textbf{0.55}/0.58 & 0.53/0.57/0.55 & \textbf{0.83}/\textbf{0.83}/0.73 \\
        \cmidrule(lr){2-7}
         & \multirowcell{2}{Context-Aware Expert}
         & Acc & 44.66 & 46.96 & 26.55 & 85.74 \\
         && Char F1 & 0.24/0.50/0.55 & --/0.54/0.57 & 0.25/0.53/0.52 & 0.23/0.65/0.63 \\
        \cmidrule(lr){2-7}
         & \multirowcell{2}{Context-Free Expert}
         & Acc & 44.19 & 44.06 & \textbf{45.24} & 76.74 \\
         && Char F1 & 0.41/0.48/\textbf{0.59} & --/0.50/\textbf{0.60} & 0.57/0.58/\textbf{0.59} & 0.81/0.82/\textbf{0.84} \\
        \bottomrule
    \end{tabular}
    \vspace{-0.3cm}
    \caption{Comparison with the performance of each expert. Char F1 scores represent the scores of \textit{few} / \textit{medium} / \textit{many} characters, respectively.}
    \vspace{-0.2cm}
    \label{Table:experiment-expert}
\end{table*}

Table~\ref{Table:experiment-english-acc} compares the baselines and our method on English datasets.
In the case-insensitive setting, although the training data we utilize are different from that of the STR baselines, our models show comparable performance.
Due to the fact that the English case-insensitive setting contains only 36 letters including the lowercase letters and numbers, the model incorporating CAFE-Net does not exhibit a significant performance improvement.
In contrast, in the case-sensitive setting, which is more challenging, our method enhances the performances even in English datasets.
These results demonstrate that CAFE-Net is more effective for languages with a large number of characters, which generally have a long-tailed distribution at the character level.
Moreover, these results indicate that the issue of long-tailed distribution at the character level is also found in English and needs to be addressed.

\begin{figure}[t!]
    \centering
    \includegraphics[width=1.0\linewidth]{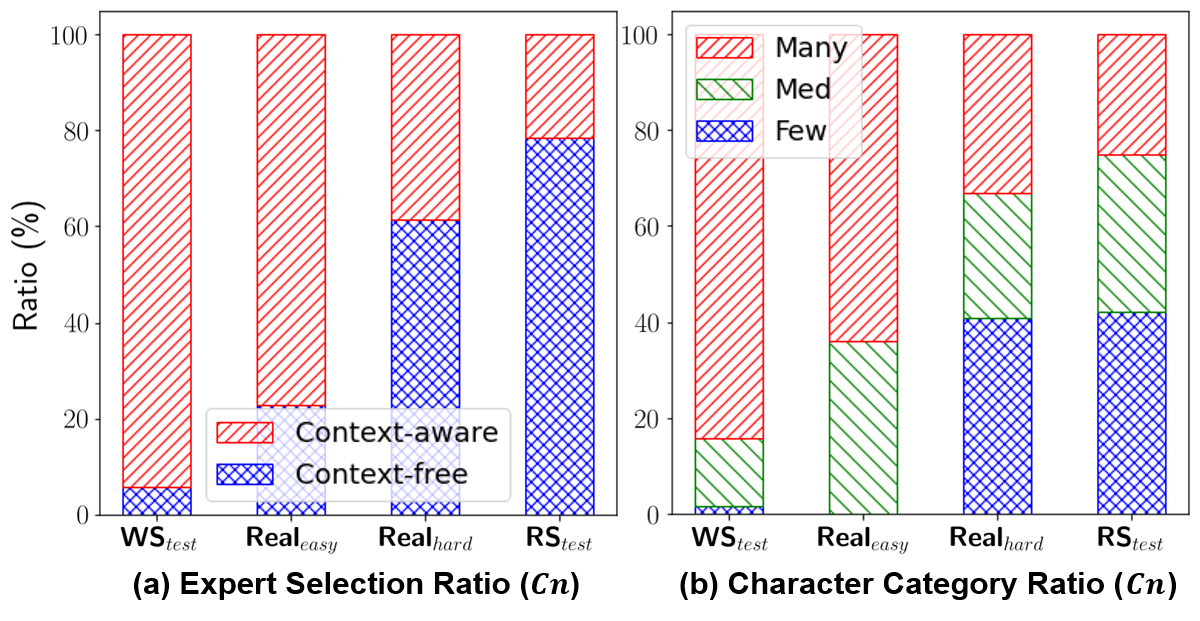}
    \vspace{-0.7cm}
    \caption{The expert selection ratio and the character ratio on Chinese. (a) We observe that CAFE-Net selects more predictions from context-aware branch for WS$_{test}$ and \textbf{Real}$_{Easy}$, while selecting more predictions from context-free branch for \textbf{Real}$_{Hard}$ and RS$_{test}$. (b) We show the proportion of \emph{many}, \emph{medium}, and \emph{few} characters in each test set.}
    \vspace{-0.3cm}
    \label{supp-fig:ratio analysis}
\end{figure}

\section*{E. Additional Results}

\noindent\textbf{Expert Selection Ratio on Chinese.}
Similar to Fig.~7 in the main paper, we also provide the relation between each expert's selection ratio and the character category ratio in the Chinese test sets.
As observed in results with Korean, we found that the ratio of \textit{many} characters is proportional to the proportion of predictions selected by the context-aware expert.
These results show that our method properly assigns the experts for different data distributions.

\noindent\textbf{Performance of Each Expert.}
Table.~\ref{Table:experiment-expert} shows the performance of each expert in CAFE-Net.
Interestingly, the accuracy of the confidence ensemble is superior to that of both context-aware and context-free experts, except for the accuracy on Korean \textbf{Synth}$_{test}$ and Chinese \textbf{Real}$_{Hard}$.
These results demonstrate that our confidence ensemble utilizes each expert properly, improving the STR performance.

\begin{figure*}[t!]
    \centering
    \includegraphics[width=1.0\linewidth]{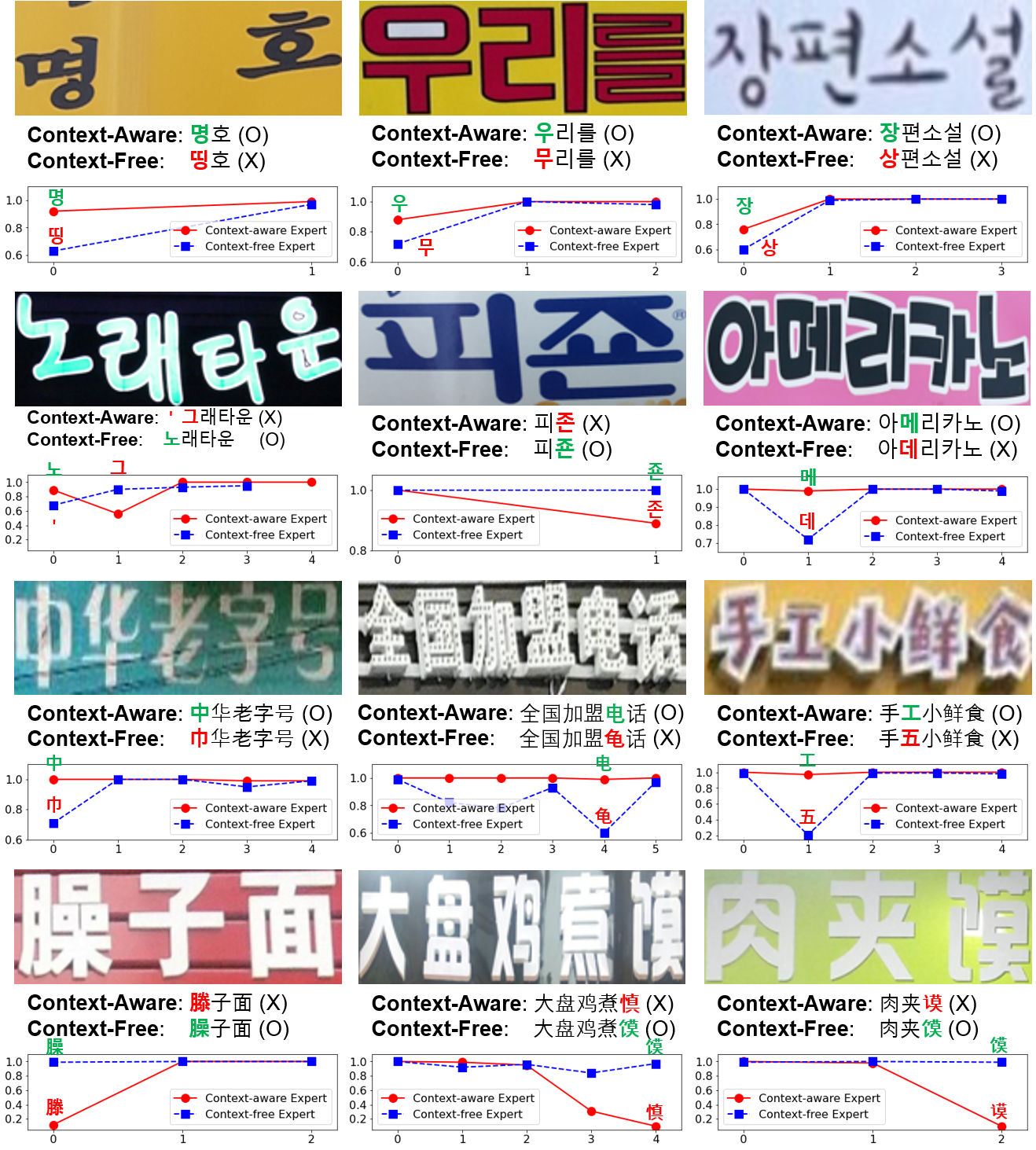}
    \caption{Additional qualitative results. For each plot under each image, the x-axis and the y-axis indicate the character sequence and the maximum softmax probability of each character, respectively.}
    \label{supp-fig:qualitative analysis}
\end{figure*}

\noindent\textbf{Additional Qualitative Results.}
We provide additional qualitative results to demonstrate that our confidence ensemble improves the STR performance by filtering the low-confident prediction of one expert and selecting the high-confident prediction of the other expert.
Fig.~\ref{supp-fig:qualitative analysis} shows the prediction results on various samples in Korean and Chinese test sets (\textit{i.e.}, \textbf{Real}).




\clearpage

\end{document}